\pdfoutput=1

\documentclass[11pt]{article}
\usepackage{float}

\usepackage[preprint]{acl}

\usepackage{times}
\usepackage{latexsym}

\usepackage[T1]{fontenc}

\usepackage[utf8]{inputenc}

\usepackage{microtype}

\usepackage{inconsolata}

\usepackage{graphicx}

\usepackage{amsmath}
\usepackage{amsfonts}
\usepackage{xspace}
\newcommand*{\imagenet}{\textsc{ImageNet1K}\xspace}
\usepackage{csquotes}
\usepackage{titletoc}

\definecolor{LightCyan}{rgb}{0.318, 0.439, 0.843}
\definecolor{LightCyan}{rgb}{0.8295, 0.85975, 0.96075}
\definecolor{IncrGreen}{rgb}{0.25, 0.55, 0.15}
\definecolor{DecrRed}{rgb}{0.55, 0.25, 0.25}
\newcommand{\qinc}[1]{\normalsize\textcolor{IncrGreen}{$\uparrow #1$}}
\newcommand{\qdec}[1]{\scriptsize\textcolor{DecrRed}{$\downarrow #1$}}
\usepackage{algorithmic}
\usepackage{algorithm}
\usepackage{comment}

\usepackage{multirow}
\usepackage{booktabs}
\usepackage{xcolor}
\usepackage{todonotes} 

\title{A Reproduction Study: The Kernel PCA Interpretation\\ of Self-Attention Fails Under Scrutiny}

\author{Karahan Sarıtaş \\
  University of Tübingen \\
  \texttt{karahan.saritas@student.uni-tuebingen.de} \\\And
  Çağatay Yıldız \\
 University of Tübingen \\
  Tübingen AI Center\\}

\begin{document}
\maketitle

\begin{abstract}

In this reproduction study, we revisit recent claims that self-attention implements kernel principal component analysis (KPCA) \cite{teo2024unveilinghiddenstructureselfattention}, positing that (i) value vectors $V$ capture the eigenvectors of the Gram matrix of the keys, and (ii) that self-attention projects queries onto the principal component axes of the key matrix $K$ in a feature space. Our analysis reveals three critical inconsistencies: (1) No alignment exists between learned self-attention value vectors and what is proposed in the KPCA perspective, with average similarity metrics (optimal cosine similarity $\leq 0.32$, linear CKA (Centered Kernel Alignment) $\leq 0.11$, kernel CKA $\leq 0.32$) indicating negligible correspondence; (2) Reported decreases in reconstruction loss $J_\text{proj}$, arguably justifying the claim that the self-attention
minimizes the projection error of KPCA, are misinterpreted, as the quantities involved differ by orders of magnitude ($\sim\!10^3$);  (3) Gram matrix eigenvalue statistics, introduced to justify that $V$ captures the eigenvector of the gram matrix, are irreproducible without undocumented implementation-specific adjustments. Across 10 transformer architectures, we conclude that the KPCA interpretation of self-attention lacks empirical support.  
\end{abstract}

\section{Introduction}

Transformers \cite{vaswani2023attentionneed, dehghani2019universaltransformers} dominate tasks spanning computer vision \cite{dosovitskiy2021imageworth16x16words, liu2021swintransformerhierarchicalvision, caron2021emergingpropertiesselfsupervisedvision, esser2021tamingtransformershighresolutionimage, parmar2018imagetransformer}, natural language processing \cite{devlin2019bertpretrainingdeepbidirectional, brown2020languagemodelsfewshotlearners, raffel2023exploringlimitstransferlearning}, and beyond \cite{chen2021decisiontransformerreinforcementlearning, huang2018musictransformer, Schwaller_2019}. At their core lies the attention mechanism, which recent works reinterpret through kernel methods \cite{tsai2019transformerdissectionunifiedunderstanding, choromanski2022rethinkingattentionperformers, chen2023primalattentionselfattentionasymmetrickernel, teo2024unveilinghiddenstructureselfattention, chowdhury2022learningtransformerkernel}. This perspective bridges transformers with classical kernel techniques, leveraging their interpretability \cite{Ponte_2017} and computational efficiency via the kernel trick \cite{vankadara2019optimalitykernelshighdimensionalclustering}. 

Recent work by \citet{teo2024unveilinghiddenstructureselfattention} reframes self-attention through the lens of kernel principal component analysis (KPCA), proposing that self-attention implicitly projects query vectors onto the principal component axes of the key matrix in a feature space. The authors further assert that the value matrix $V$ converges to encode the eigenvectors of the Gram matrix formed by the key vectors. While theoretical proofs for such convergence under stochastic gradient descent training remain challenging due to non-convex optimization dynamics, they provide empirical justifications for their claims. This theory, \textit{if empirically validated}, offers significant potential to enhance the interpretability and efficiency of state-of-the-art methods in Computer Vision, NLP, and related domains. By reducing the quadratic complexity of transformers through scalable kernel methods \cite{choromanski2022rethinkingattentionperformers}, it can unlock practical improvements in resource-intensive applications.  

In this reproduction study, we empirically validate the core claims of the KPCA interpretation proposed by \citet{teo2024unveilinghiddenstructureselfattention} Our findings challenge the validity of the KPCA analogy, revealing inconsistencies the empirical justifications proposed that question the robustness of the original claims. Specifically, we evaluate (1) the correspondence between attention-learned value vectors and the KPCA correspondence, (2) reconstruction loss and its true interpretation, and (3) the eigenvalue justification of proposed KPCA framework. 
Further analysis indicates that key visualizations in the prior work relied on misleading log-scale representations and non-reproducible inconsistent results, suggesting their conclusions may not hold under rigorous empirical scrutiny.

\section{A Quick Overview: Kernel PCA Analysis of Attention}

{\it \underline{Self-Attention}:} For input $X \in \mathbb{R}^{N \times d}$ (sequence length $N$, embedding dim. $d$), compute:
\begin{align}
Q = XW_Q^\top,\quad K = XW_K^\top,\quad V = XW_V^\top
\end{align}
with weight matrices $W_Q,W_K \in \mathbb{R}^{d_q \times d}$, $W_V \in \mathbb{R}^{d_v \times d}$. Let $q_i := Q[i,:]$, $k_i := K[i,:]$, and $v_i := V[i,:]$ denote the query/key/value vectors for position $i$ (row vectors). The output $h_i$ is then:
\begin{align}
h_i = \sum_{j=1}^N \underbrace{\sigma\left(\frac{q_i K^\top}{\sqrt{d_q}}\right)_j}_{\text{attention weight} \;\alpha_{ij}} v_j, \quad
\sigma(z)_i = \frac{e^{z_i}}{\sum_{j=1}^N e^{z_j}}
\end{align}
where $\sigma$ applies row-wise softmax normalization to the scaled attention score matrix $QK^\top / \sqrt{d_q}$. Output vector $h_i \in \mathbb{R}^{d_v}$ is the convex combination of value vectors $v_j$, weighted by $\alpha_{ij}$. \\
{\it \underline{Kernel PCA Derivation}:} Let $\{k_j\}_{j=1}^N \subset \mathbb{R}^{d_q}$ be mapped through $\varphi(k_j) := \phi(k_j)/g(k_j)$ and scaling $g(k_j) = \sum_{j'}k(k_j,k_{j'})$. Centered key features $\tilde{\varphi}(k_j) = \varphi(k_j) - \frac{1}{N}\sum_{j'}\varphi(k_{j'})$ yield covariance:
\begin{align}
\label{eq:C}
C = \tfrac{1}{N}\sum_j \tilde{\varphi}(k_j)\tilde{\varphi}(k_j)^\top
\end{align}
Eigenvectors of $C$ are denoted by $u_d$ with eigenvalue $\lambda_d$, which can be expressed as a weighted sum of the keys $
u_d = \sum_{j=1}^N a_{dj}\tilde{\varphi}(k_j) $. Weights $a_{dj}$ are given by $a_{dj} = \tfrac{1}{N\lambda_d}\tilde{\varphi}(k_j)^\top u_d$.
Then the kernel is set $k(x,y) =\exp\left(x^\top y / \sqrt{d_q}\right)$ to resemble the scaled softmax attention. Projection score $h_{id}$ ($d^\text{th}$ entry of the output vector $h_i \in \mathbb{R}^{d_v}$) of query $q_i$ onto principal component $u_d$ yields:
\begin{align}
\label{eqn:derive_h}
h_{id} &= \varphi(q_i)^{\top}u_d \nonumber 
= \sum_{j=1}^N \frac{k(q_i, k_j)}{g(q_i)}\dot{v}_{jd}
\end{align}
where $\dot{v}_{jd} := \frac{a_{dj}}{g(k_j)} - \frac{1}{N}\sum_{j'=1}^N\frac{a_{dj'}}{g(k_{j})}$.
Here comes one of the main claims of the paper, which suggests that the \textbf{self-attention learned value vectors $v_j = W_Vx_j$ converge to the KPCA term $\dot{v}_{j}$} during training (see Section 2.2 in \citet {teo2024unveilinghiddenstructureselfattention}), and therefore concluding that attention outputs are projections of the query vectors onto the principal components axes of the key matrix $K$ in a feature space $\varphi(\cdot)$. 

To determine coefficients $\{a_{dj}\}$, they define the centered Gram matrix $\tilde{K}_\varphi \in \mathbb{R}^{N \times N}$ where $\tilde{K}_\varphi(i,j) = \tilde{\varphi}(k_i)^\top\tilde{\varphi}(k_j)$, which can be calculated during the forward pass using key values. Substituting the eigenvector expansion $u_d = \sum_j a_{dj}\tilde{\varphi}(k_j)$ into $Cu_d = \lambda_du_d$ gives:  
\begin{align}  
\frac{1}{N}\sum_{j=1}^N \tilde{\varphi}(k_j)\tilde{\varphi}(k_j)^\top\sum_{j'=1}^N a_{dj'}\tilde{\varphi}(k_{j'})&= \lambda_d \sum_{j=1}^N a_{dj}\tilde{\varphi}(k_j)  
\end{align}  
Left-multiplying by $\tilde{\varphi}(k_i)^\top$ yields:  
\begin{align}  
\tilde{K}_\varphi^2 a_d = \lambda_d N \tilde{K}_\varphi a_d \implies \tilde{K}_\varphi a_d = \lambda_d N a_d,  
\end{align}  
where $a_d = [a_{d1},...,a_{dN}]^\top$ are eigenvectors of $\tilde{K}_\varphi$. Defining $G := \text{diag}(\frac{1}{g(k_1)},..., \frac{1}{g(k_N)})$, $\textbf{1}_N \in \mathbb{R}^{N \times N}$ consisting of $\frac{1}{N}$ in all entries, and $A := [a_1, ..., a_{d_v}] \in \mathbb{R}^{N \times d_v}$ consisting of $d_v$ eigenvectors of the gram matrix, KPCA value matrix $\dot{V}_\text{KPCA} = [\dot{v}_1, ..., \dot{v}_N]^\top \in \mathbb{R}^{N \times d_v}$ can be expressed as follows: 
\begin{align}
\label{eqV}
\dot{V}_\text{KPCA} &= GA - G \textbf{1}_NA \\ \implies  &\hat{a}_d = (I - \textbf{1}_N)^{-1} G^{-1} V[:,d] 
 \end{align}

Building on the hypothesis that self-attention’s learned value vectors $V$ converge to kernel PCA coefficients $\dot{V}_\text{KPCA}$, \citet{teo2024unveilinghiddenstructureselfattention} assert that the value matrix encodes the eigenvectors of the Gram matrix derived from key vectors in a feature space. In Section~\ref{sec:experiments}, we empirically test their claims by analyzing their proposed evidence for eigenvector alignment and projection error minimization.

\section{Experiments}
\label{sec:experiments}

\paragraph{Is self-attention learned $V \approx \dot{V}_\text{KPCA}$?}

We first assess whether attention-learned value matrices $V$ align with theoretical kernel PCA counterparts $\dot{V}$, evaluating 10 vision transformers: 6 DeiT models (\texttt{tiny}, \texttt{small}, \texttt{base}, and their distilled variants (patch $16$)) \cite{touvron2021trainingdataefficientimagetransformers} and 4 ViT variants (\texttt{tiny}/\texttt{small}/\texttt{base}/\texttt{large}) \cite{dosovitskiy2021imageworth16x16words}, all trained on \imagenet \cite{russakovsky2015imagenetlargescalevisual} with image size $224 \times 224$. We analyze each attention head in each layer using a random selection of $100$ images during inference. We calculate $\dot{V}_{\text{KPCA}}$ using Equation~\ref{eqV}, where we first calculate the Gram matrix $K_\varphi$, center it, and then extract its eigenvectors to achieve the matrix $A$. We use the top $d_v$ eigenvectors of $A$ to construct $\dot{V}_\text{KPCA}$.

 We first compare matrix entries pairwise, checking if \( |\text{input}_i - \text{other}_i| \leq 10^{-3} + 10^{-5} \times |\text{other}_i| \), using relatively higher error thresholds to avoid false negatives. Across all combinations of model $\times$ image $\times$ layer $\times$ head, we conduct 114,000 tests, none of which passes the check. As this criterion may be overly stringent, we proceed with the following relaxed approaches.

 We compute cosine similarity between self-attention and KPCA value vectors. To satisfy $V \approx \dot{V}_\text{KPCA}$, we compare:  
(1) direct column-wise matches, and  
(2) optimal column alignment via \texttt{scipy}'s Jonker-Volgenant algorithm \cite{7738348} implementation using cosine distance costs to test if the hypothesis holds in the best scenario possible. Then, as a final approach to measure matrix similarity, we employ Centered Kernel Alignment (CKA) \cite{kornblith2019similarityneuralnetworkrepresentations} - which was originally used to measure the similarity of neural network representations. All comparisons are conducted after normalizing vectors remove the sensitivity to vector magnitudes.  

As illustrated in Table~\ref{tab:similarity_measures}, all four similarity measures yield relatively low values across the examined models, failing to provide compelling evidence that $V \approx \dot{V}_{\text{KPCA}}$ at the conclusion of training. Even the most promising metric—Maximum Optimal Cosine Similarity with Jonker-Volgenant matching—reaches only 0.32 at its peak, suggesting limited alignment between the attention-learned value matrices and their theoretically proposed counterparts.

\begin{table}[ht]
    \centering
    \caption{Similarity results between attention-learned value matrix $V$ and proposed $\dot{V}_\text{KPCA}$ using the following metrics: MDC: Max Direct Cosine Similarity, MOC: Max Optimal Cosine Similarity using Jonker-Volgenant matching, LCKA: Linear CKA, KCKA: Kernel CKA}
    \label{tab:similarity_measures}
    \scalebox{0.85}{\begin{tabular}{l|c|c|c|c}
\hline
\hline
    \multirow{2}{*}{\;\;\;\;\;\textbf{Model}} & \multicolumn{4}{c}{\textbf{Similarity Measures}} \\ \cline{2-5} \\[-0.9em]
    & MDC & MOC & LCKA & KCKA \\
    \midrule
    ViT-Tiny & 0.09 & 0.29 & 0.06 & 0.28 \\
    ViT-Small & 0.11 & 0.30 & 0.05 & 0.27 \\
    ViT-Base & 0.14 & 0.30 & 0.06 & 0.28 \\
    ViT-Large & 0.13 & 0.30 & 0.06 & 0.25 \\
    DeiT-Tiny & 0.15 & 0.31 & 0.11 & 0.31 \\
    DeiT-Small & 0.11 & 0.31 & 0.08 & 0.28 \\
    DeiT-Base & 0.12 & 0.32 & 0.10 & 0.29 \\
    DeiT-Tiny-D & 0.11 & 0.31 & 0.11 & 0.32 \\
    DeiT-Small-D & 0.11 & 0.31 & 0.09 & 0.29 \\
    DeiT-Base-D & 0.11 & 0.32 & 0.10 & 0.28 \\
\hline
\hline
    \end{tabular}}
\end{table}

Having found no evidence that self-attention-learned $V$ matrices converge to KPCA theoretical values, we now analyze the authors’ empirical justifications for their hypothesis.

\paragraph{Does the decrease in $J_\text{proj}$ imply convergence?}
We reproduce the projection error minimization plot from \cite{teo2024unveilinghiddenstructureselfattention}, where the error is defined as:  
\[  
J_{\text{proj}} = \frac{1}{N} \sum_{i=1}^{N} \left\| \varphi(q_i) - \sum_{d=1}^{d_v} h_{id} u_d \right\|^2 
\]  
While our implementation replicates the numerical results using the authors’ code\footnote{\url{https://github.com/rachtsy/KPCA_code/blob/07e579a/Reconstruction/softmax.py\#L73}}, critical discrepancies arise in implementation. The original work visualizes $\log(J_{\text{proj}})$ without explicitly stating this logarithmic scaling in their manuscript, obscuring the raw magnitude of the projection error. Furthermore, the omission of a $\sqrt{d_v}$ scaling factor for $\varphi(q_i)$ leads to inflated $\|\varphi(q_i)\|^2$ values resulting in values of $e^{35}$ even after 300-epoch training. We train both ViT-Tiny and DeiT-Tiny on \imagenet, and plot the minimization error in Figure~\ref{fig:min_error} after correcting the normalization and adopting mean absolute error (see Appendix~\ref{appx:loss_details}).

\begin{figure}[ht]
    \centering
    \includegraphics[width=1\linewidth]{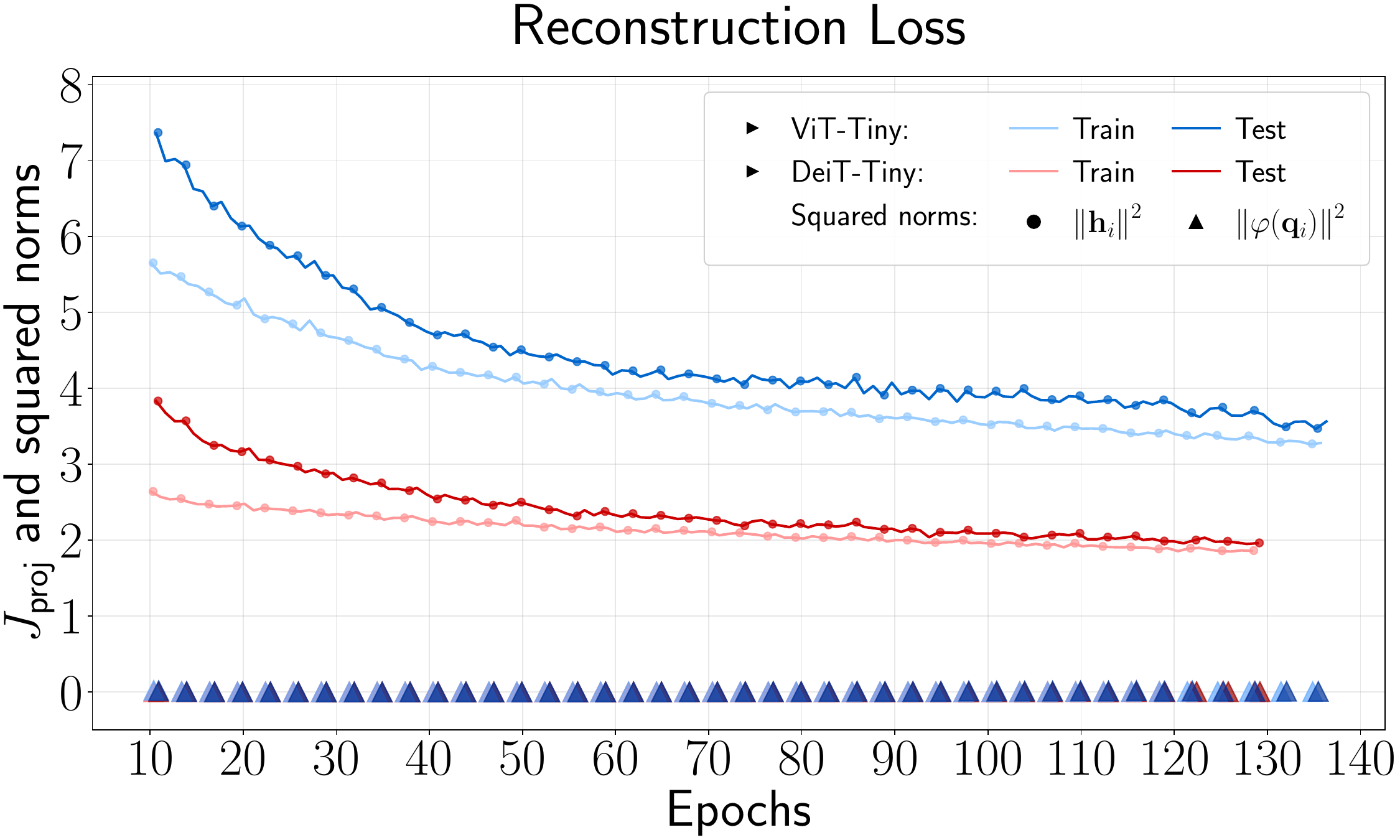}
    \caption{Reconstruction loss ($J_{\text{proj}}$) over training epochs for ViT-Tiny and DeiT-Tiny models, along with the values of the individual squared norms, shown with markers.  Circle markers indicate average of squared output norms ($\|\mathbf{h}_i\|^2$) and triangle markers (extremely low values around $10^{-3}$) show the average of squared feature map norms ($\|\varphi(\mathbf{q}_i)\|^2$).}
    \label{fig:min_error}
\end{figure}

At first, decreasing projection loss $J_\text{proj}$ may seem to indicate a meaningful alignment between the quantities; however, analysis of individual squared norms reveals a more nuanced picture. As shown by the markers, $\|\varphi(q_i)\|^2$ values (around $10^{-3}$) remain orders of magnitude smaller than $\|h_i\|^2$ throughout training. In practice, the observed error reduction stems predominantly from decreasing $\|h_i\|^2$ magnitudes rather than genuine convergence between $\varphi(q_i)$ and the reconstruction. Our observations on vision transformers generalize to the language models with transformers (See Appendix~\ref{app:proj_vis} for additional visualizations). 

\paragraph{Do eigenvalues of $\tilde{K}_\varphi$ match with the reported results?}

The authors empirically verified the relationship  
$\frac{\tilde{K}_{\varphi} \hat{a}_d}{N \hat{a}_d} = \gamma = [\gamma_1, \ldots, \gamma_N], \quad$ where $ \gamma_1 = \cdots = \gamma_N = \text{constant},$  
which they interpret as confirmation that $\hat{a}_d$ is an eigenvector of $\tilde{K}_{\varphi}$ (with eigenvalue $N\gamma$).

Plots of the means and standard deviations of absolute differences $|\gamma_i - \gamma_j|$ in the vector $\mathbf{1}\lambda_d$ can be misleading, as small values may yield low differences without satisfying the eigenvalue constraint (Appendix~\ref{app:eigenvalue}). Therefore we have to focus on reproducing the actual eigenvalues. The authors emphasize that the eigenvalues' magnitudes---averaged across all attention heads and layers---are substantially larger, with maximum, minimum, mean, and median values of 648.46, 4.65, 40.07, and 17.73, respectively, far exceeding $|\gamma_i - \gamma_j|$. Unfortunately, they provide no reproducible implementation for this claim. Our analysis of eigenvalues of $\tilde{K}_{\varphi}$ across 10 distinct transformer models demonstrates fundamental inconsistencies: the empirical eigenvalue distribution directly contradicts the reported values to justify their claims. We compute absolute eigenvalues across all attention heads and layers for each image, average them \textit{by eigenvalue rank} (see Appendix~\ref{app:eig_stats}), then derive per-image statistics (max/min/mean/median) from these rank-wise averages. We report mean ± standard deviation over 25 randomly selected \imagenet images.

\begin{table}[htbp]
\centering
\caption{Eigenvalue Statistics for Vision Transformer Models ($\times 10^{-6}$)}
\label{tab:eigenvalues}
\scalebox{0.85}{
\begin{tabular}{@{}l|c|c|c|c@{}}
\hline
\hline
    \multirow{2}{*}{\;\;\;\;\;\textbf{Model}} & \multicolumn{4}{c}{\textbf{ Eigenvalue Statistics}} \\ \cline{2-5} \\[-0.9em]
    & Max & Min & Mean & Median \\
    \midrule
ViT-Tiny & $147 \pm 11$ & $17 \pm 5$ & $37 \pm 7$ & $30 \pm 7$ \\
ViT-Small & $181 \pm 22$ & $17 \pm 4$ & $36 \pm 6$ & $28 \pm 5$ \\
ViT-Base & $206 \pm 30$ & $15 \pm 4$ & $33 \pm 6$ & $25 \pm 5$ \\
ViT-Large & $177 \pm 22$ & $21 \pm 5$ & $42 \pm 6$ & $34 \pm 6$ \\
DeiT-Tiny & $325 \pm 5$ & $34 \pm 10$ & $65 \pm 10$ & $53 \pm 11$ \\
DeiT-Small & $306 \pm 4$ & $34 \pm 9$ & $66 \pm 11$ & $54 \pm 11$ \\
DeiT-Base & $259 \pm 7$ & $35 \pm 9$ & $64 \pm 10$ & $54 \pm 10$ \\
DeiT-Tiny-D & $205 \pm 7$ & $32 \pm 9$ & $61 \pm 10$ & $51 \pm 10$ \\
DeiT-Small-D & $224 \pm 7$ & $33 \pm 9$ & $63 \pm 10$ & $53 \pm 10$ \\
DeiT-Base-D & $226 \pm 6$ & $36 \pm 9$ & $67 \pm 10$ & $56 \pm 10$ \\
\hline
\hline
\multicolumn{5}{@{}l@{}}{}\\
\end{tabular}}
\end{table}

Table~\ref{tab:eigenvalues} reveals eigenvalues of $\tilde{K}_{\varphi}$ on the order of $10^{-6}$—orders of magnitude smaller than those reported in \cite{teo2024unveilinghiddenstructureselfattention}. This discrepancy not only challenges the reproducibility of their spectral analysis but also undermines the validity of the $\gamma$-difference plots to validate self-attention’s convergence to KPCA value vectors.

\section{Conclusion}
\label{sec:conclusion}

In essence, the kernel PCA interpretation of self-attention proposed by \citet{teo2024unveilinghiddenstructureselfattention} lacks empirical and theoretical robustness under detailed scrutiny. Our results extend to language models: the similarity between $V$ and $\dot{V}_\text{KPCA}$ stays low, and the two norms diverge (see Appendix~\ref{sec:language_models}). We emphasize that this critique neither disputes the viability of Robust PCA (RPCA) as an algorithm nor asserts that the self-attention cannot be interpreted as a projection—rather, it challenges the proposed framework’s empirical and theoretical foundations. Specifically, the claim that the self-attention can be derived from kernel PCA (and therefore can be replaced with) by the proposed mechanism, is unsupported by reproducible evidence. We believe that the RPCA’s improvements stem from its complementary role within the existing architecture, using the symmetric self-attention mechanism as a low-rank approximator in its Principal Component Pursuit (PCP) algorithm rather than replacing it outright. To ensure reproducibility during the peer review as well, we provide anonymized code\footnote{\url{https://anonymous.4open.science/r/Reproduction-Study-KPCA-B01F/}}.

The interpretation of self-attention has become a rapidly developing area, with numerous works proposing formulations from different mathematical perspectives \cite{chen2023primalattentionselfattentionasymmetrickernel, choromanski2022rethinkingattentionperformers, tsai2019transformerdissectionunifiedunderstanding, nguyen2024primaldualframeworktransformersneural}. However, such rapid progress risks false positives in research community. We hope our work helps researchers navigate this landscape more efficiently, focusing attention on evidence-based progress rather than superficially consistent narratives, mis-interpreted plots or undocumented, unconventional implementation practices. While the interpretation of self-attention mechanisms as projections of input, key, or query vectors remains an open research question, our empirical evidence directly refutes how this mechanism is characterized in \cite{teo2024unveilinghiddenstructureselfattention}.  

\break 
\break

\section{Limitations}


Despite our extensive evaluation, several practical limitations should be acknowledged. First, we had to resort to a proxy reconstruction loss (similar to the original work \cite{teo2024unveilinghiddenstructureselfattention} (e.g., MAE over squared norm differences) rather than an exhaustive permutation-based matching (see Appendix~\ref{app:sensitive}). Secondly, the numerical instability in computing the eigenvalues of the centered Gram matrix $\tilde{K}_\varphi$ forced us to adopt pre-processing steps ($Z$-score normalization) that, although minimally impacting overall trends and conclusions, produces different eigenvalues. Lastly, we can compare the self-attention learned $V$ with the KPCA counterpart $\dot{V}_{\text{KPCA}}$ through two directions:  
First, estimating eigenvectors $\hat{A} = (I - \mathbf{1}_N)^{-1} G^{-1} V$ to verify alignment with $\tilde{K}_{\varphi}$'s eigenvectors, but this approach is not feasible due to the singular centering matrix $I - \mathbf{1}_N$, which introduces numerical instability during inversion.  
Alternatively, we compute $A$ directly from the eigenvectors of $\tilde{K}_{\varphi}$ and validate whether $G A - G \mathbf{1}_N A \approx V$ holds.  
Due to the numerical instability in the first method, we adopt the second approach in our analysis


\bibliography{custom}

\newpage 

\appendix

\begin{center}
{\bf {Supplement to \enquote{A Reproduction Study: The Kernel PCA Interpretation
of Self-Attention Fails Under Scrutiny}}}
\end{center}

\section{Calculation of \texorpdfstring{$J_\text{proj}$}{Jproj} and Practical Limitations}
\label{appx:loss_details}

Main claim by \citet{teo2024unveilinghiddenstructureselfattention} is that the output $h_i \in \mathbb{R}^{d_v}$ of self-attention is equivalent to the projection of query vector $q_i$ onto the principal components of the key matrix $K \in \mathbb{R}^{N \times N}$ in a feature space $\varphi(\cdot)$. Projection scores can be expressed as $h_{id} = \varphi(q_i)^\top u_d$, where $u_d$ is an eigenvector of the matrix $C$ (Equation \ref{eq:C}). If $u_d$ is a unit eigenvector, then it is a \textit{normalized} projection score, otherwise \textit{unnormalized} which requires dividing by the scalar $u_d^\top u_d$ to normalize it.

To reconstruct the projected vector, we sum the projection scores along each principal component: ${\hat{\varphi}(q_i)} = \sum_{d=1}^{d_v} h_{id}u_d = \sum_{d=1}^{d_v} (u_d^\top\varphi(q_i))u_d$, which gives us the reconstructed vector in the original embedding space.
\begin{figure*}
\begin{align*}
    J_{\text{proj}} &= \frac{1}{N} \sum_{i=1}^{N} \left\| \varphi(q_i) - \sum_{d=1}^{d_v} h_{id} u_d \right\|^2 \\
    &= \frac{1}{N} \sum_{i=1}^{N} \left(\varphi(q_i) - \sum_d^{d_v}  h_{id} u_d\right)^\top 
    \left(\varphi(q_i) - \sum_d^{d_v}  h_{id} u_d \right) \\
    &= \frac{1}{N} \sum_{i=1}^{N} \left( \varphi(q_i)^\top \varphi(q_i) -  \sum_d^{d_v} h_{id}\varphi(q_i)^\top   u_d - \sum_d^{d_v}  h_{id} u_d^\top \varphi(q_i) 
    + \sum_m^{d_v} \sum_n^{d_v} u_m^\top  u_n h_{ia} h_{ib} \right) \\
      &= \frac{1}{N} \sum_{i=1}^{N} \left( \left\| \varphi(q_i) \right\|^2 -  \underbrace{\sum_d^{d_v} h_{id}^2}_{\left\| h_i \right\|^2} - \underbrace{\sum_d^{d_v} h_{id}^2}_{\left\| h_i \right\|^2}
    +\sum_m^{d_v} \sum_n^{d_v} u_m^\top  u_n h_{ia} h_{ib} \right) \\
     &= \frac{1}{N} \sum_{i=1}^{N} \left( \left\| \varphi(q_i) \right\|^2 -  2 \left\| h_i\right\|^2
    + \underbrace{\sum_m^{d_v} \sum_n^{d_v} u_m^\top  u_n h_{ia} h_{ib}}_{\left\| h_i\right\|^2 \; \text{if orthonormal eigenvectors}} \right) \\
\end{align*}
\label{fig:equation}
\end{figure*}

If the eigenvectors are orthonormal (unit and orthogonal to each other), then the last equation reduces to the following squared norm difference (using $u_a^\top u_b = 0 \;\text{if}\; a \neq b, \;\text{otherwise}\; 1$): $\frac{1}{N} \sum_{i=1}^{N} \left( \left\| \varphi(q_i) \right\|^2 - \left\| h_i \right\|^2 \right)$. A very useful property of this equation is that it is an \enquote{eigenvector-invariant} computation, meaning we don't need to compute individual eigenvectors or assign them to corresponding rows of the output matrix $H \in \mathbb{R}^{N \times d_v}$. If eigenvectors aren't orthonormal, we must use the original equation for correct loss calculation. However, this introduces a technical challenge: \textit{if the theory holds}, we do know each component $h_{id}$ of output vector $h_i \in \mathbb{R}^{d_v}$ represents the projection score along eigenvector $u_d$ -- but we do not know which eigenvector of $C$ corresponds to $u_d$. For $d_v = 64$, the combinatorial permutation alignment problem between $d_v$ eigenvectors and components exhibits factorial computational complexity $\mathcal{O}(d_v!)$, fundamentally limiting practical verification. Due to this computational bottleneck, we used the squared norm difference (as in original work) to maintain eigenvector-invariant computation. However, $\left\| \varphi(q_i) \right\|^2 \geq \left\| h_i \right\|^2$ is not guaranteed without orthonormality, so we switched to Mean Absolute Error:

\begin{equation*}
    J_{\text{proj}} = \frac{1}{N} \sum_{i=1}^{N} \left| \left\| \varphi(q_i) \right\|^2 - \left\| h_i \right\|^2 \right|
\end{equation*}

\subsection{Eigenvector Assignment Sensitivity in Projection Loss}
\label{app:sensitive}

With a toy example, we demonstrate that different selections of eigenvectors $\{u_d\}_{d=1}^{d_v}$ yield different projection loss values. While the first two terms in the projection loss, $\left\| \varphi(q_i) \right\|^2$ and $\left\| h_i\right\|^2$, are invariant to eigenvector selection, the critical cross-term $\sum_{m=1}^{d_v} \sum_{n=1}^{d_v} u_m^\top u_n h_{im} h_{in}$ exhibits high sensitivity to the specific assignment of eigenvectors.

Consider a minimal example where an output representation $h_i = [1, 2]$ is projected into a 2-dimensional space ($d_v = 2$). Given two eigenvectors $[1, 1]^\top$ and $[-1, 0]^\top$, cross-term can be evaluated for two different assignments.
Under assignment $A_1: u_1 = [1, 1]^\top, u_2 = [-1, 0]^\top$, we obtain $= 2 - 2 - 2 + 4 = 2$, whereas under assignment $A_2: u_1 = [-1, 0]^\top, u_2 = [1, 1]^\top$, we obtain $1 - 2 - 2 + 8 = 5$, resulting in different loss values.

Simply permuting the assignment of identical eigenvectors can yield substantially different loss values. To compute the actual loss, would need to evaluate $d_v!$ different assignment permutations to identify the optimal configuration—rendering the approach computationally prohibitive. To avoid the excessive computation, we adopt the proxy MAE loss, which eliminates this assignment sensitivity.

\subsection{Additional Details on Projection Error Minimization}
\label{app:proj_vis}

We evaluate the projection error $J_\text{proj}$ for ViT-Tiny and DeiT-Tiny models. Tollowing the methodology of \cite{teo2024unveilinghiddenstructureselfattention}, the reconstruction loss is computed on the same batch of images coming from the training set, with results averaged across layers, attention heads, and batches to align with the original implementation.  

\begin{figure*}[t]
    \centering Distribution of
    $\left\| \varphi(q_i) \right\|^2$ (blue) and $\left\| h_i \right\|^2$ (red) across transformer layers (log-scale)
    \begin{tabular}{@{}c@{\hspace{0.02\textwidth}}c@{\hspace{0.02\textwidth}}c@{}}
        \includegraphics[width=0.33\textwidth]{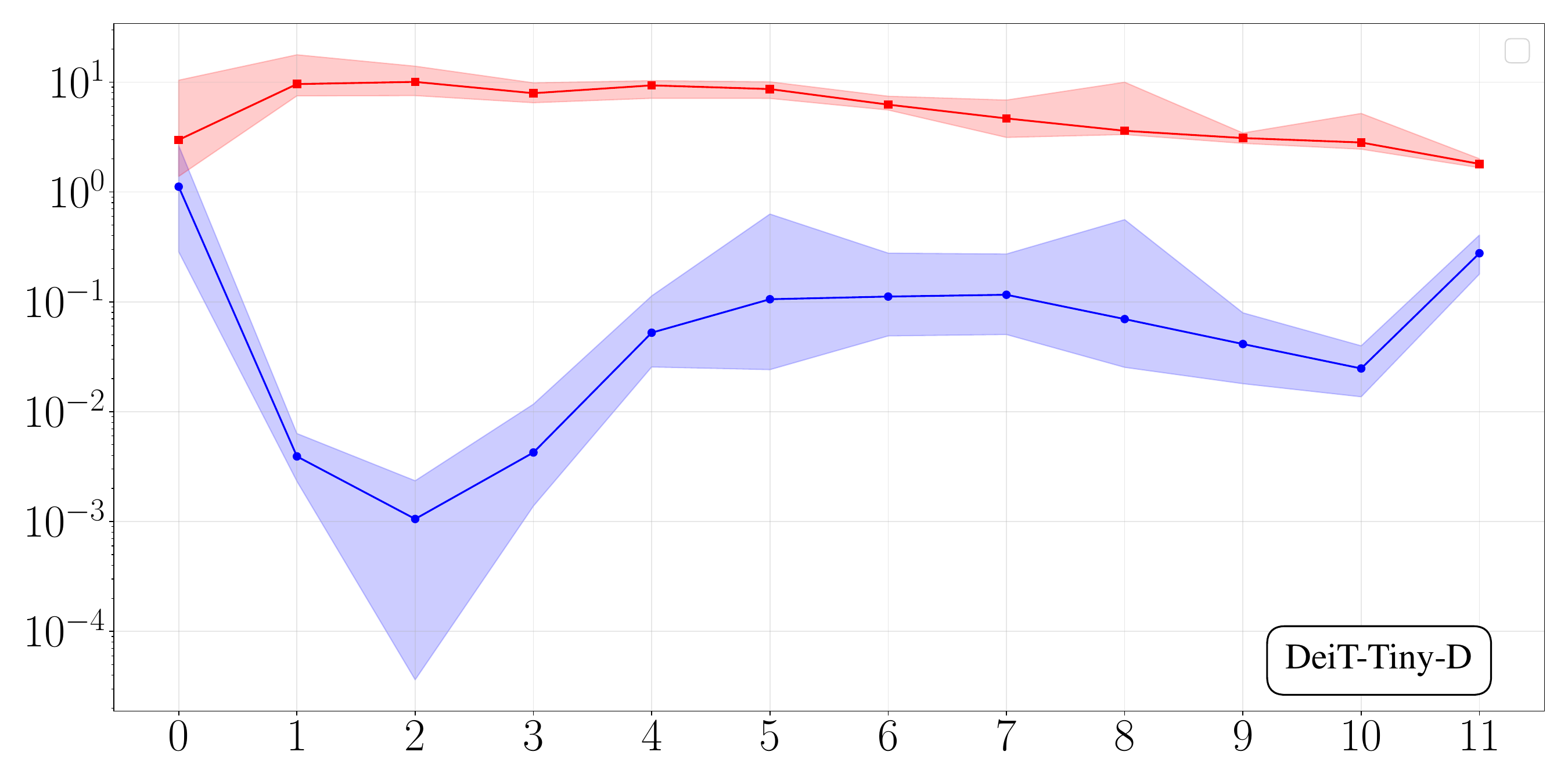} &
        \includegraphics[width=0.33\textwidth]{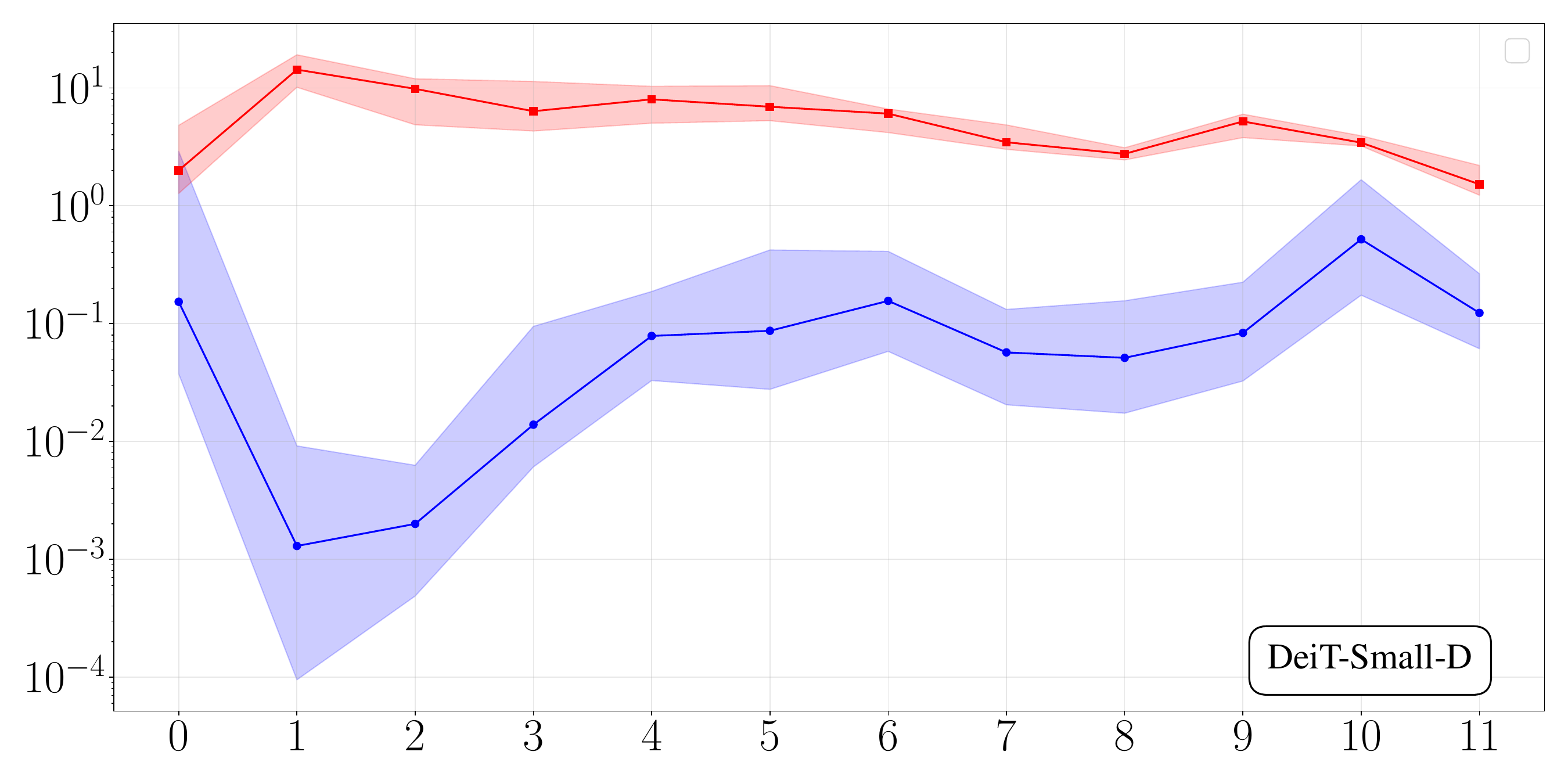} &
        \includegraphics[width=0.33\textwidth]{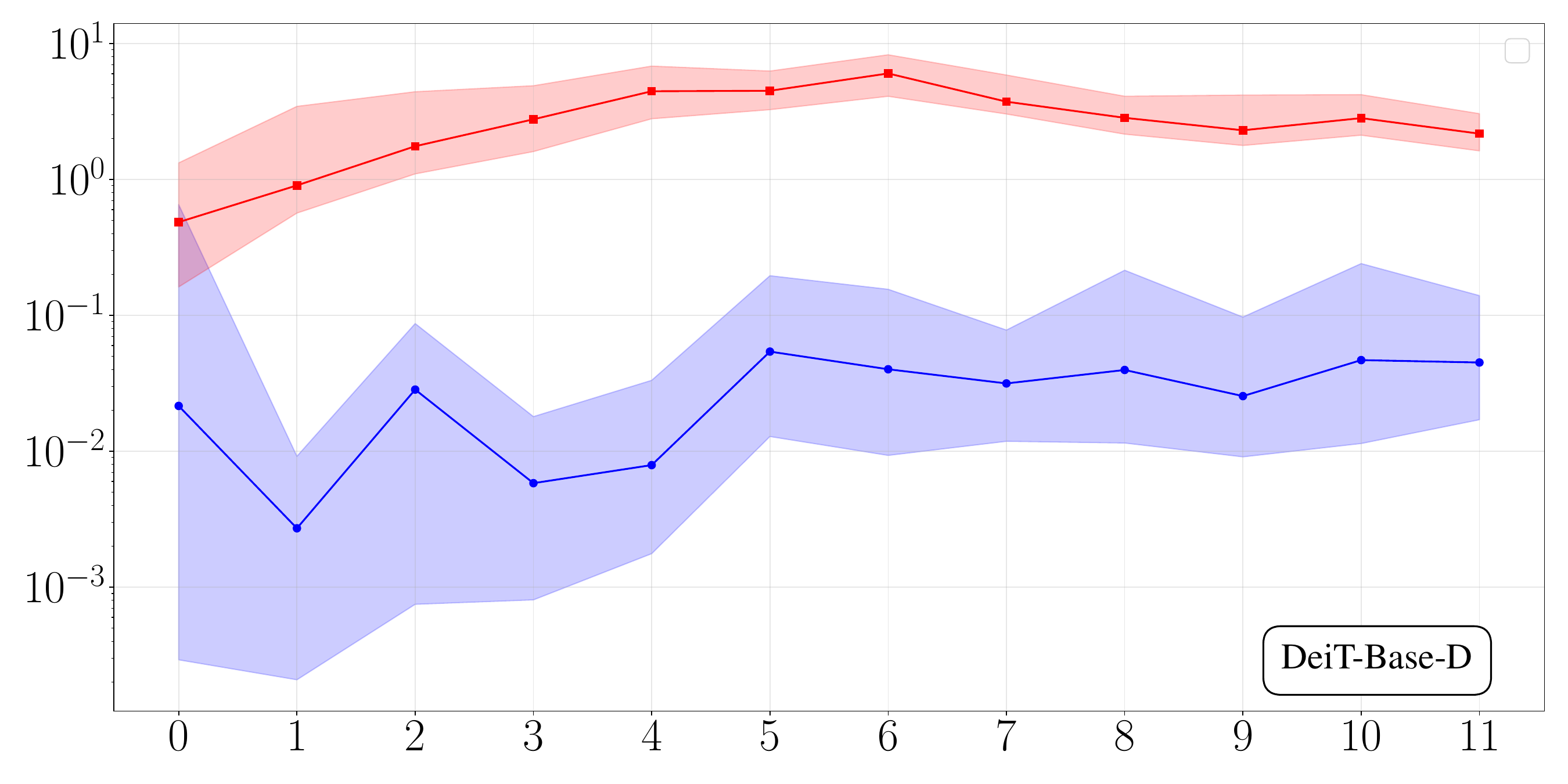} \\
        \includegraphics[width=0.33\textwidth]{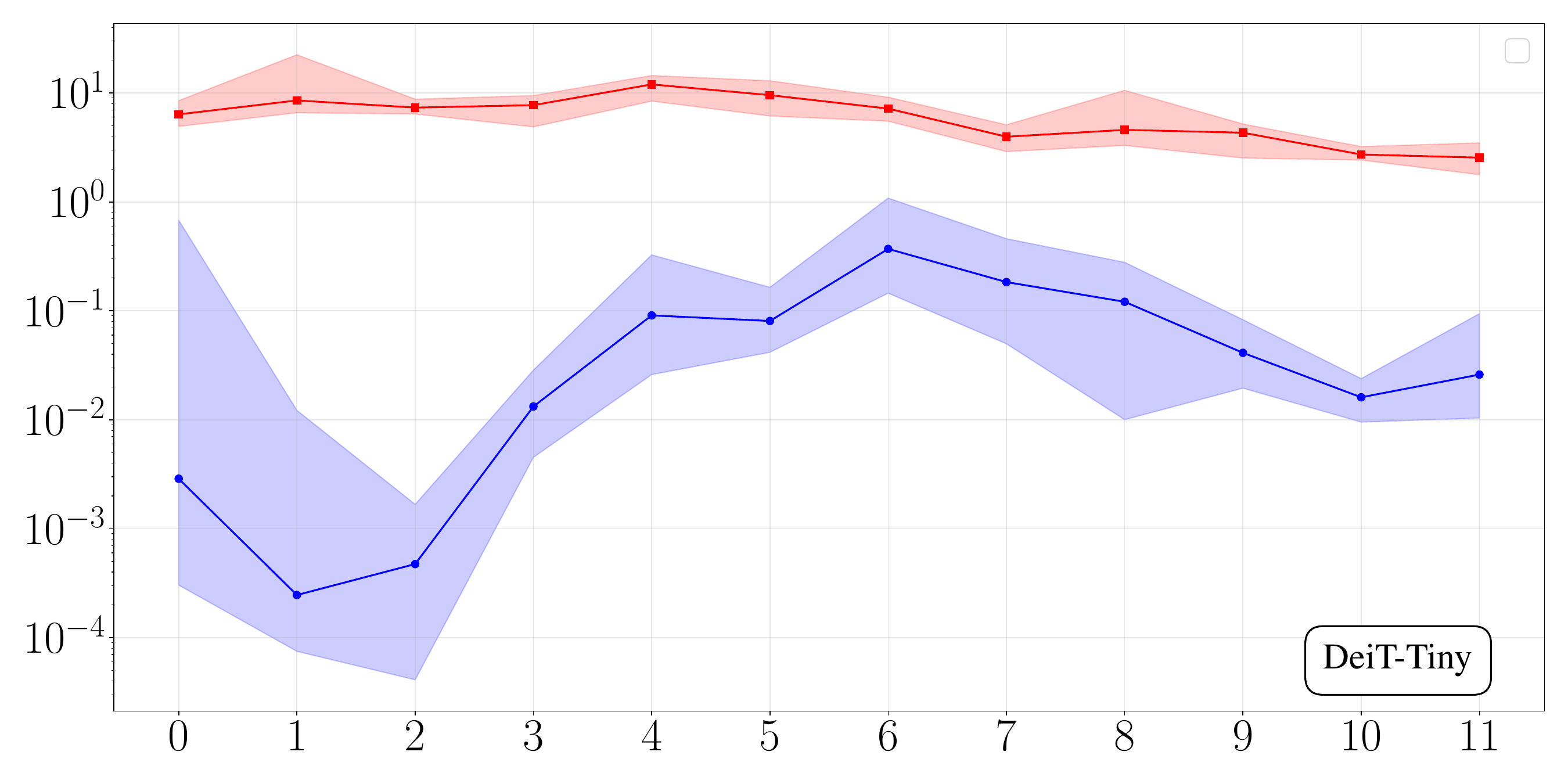} &
        \includegraphics[width=0.33\textwidth]{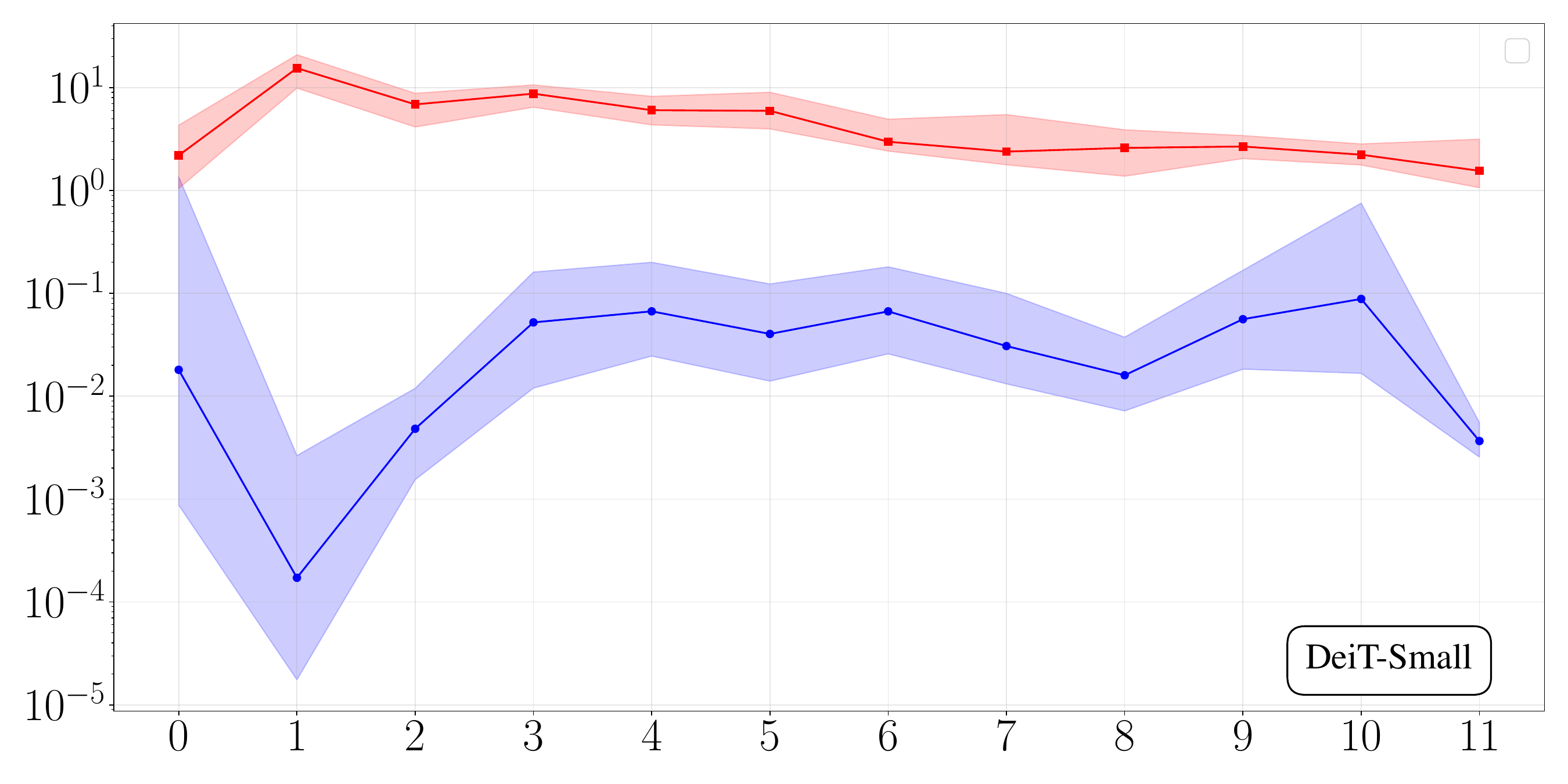} &
        \includegraphics[width=0.33\textwidth]{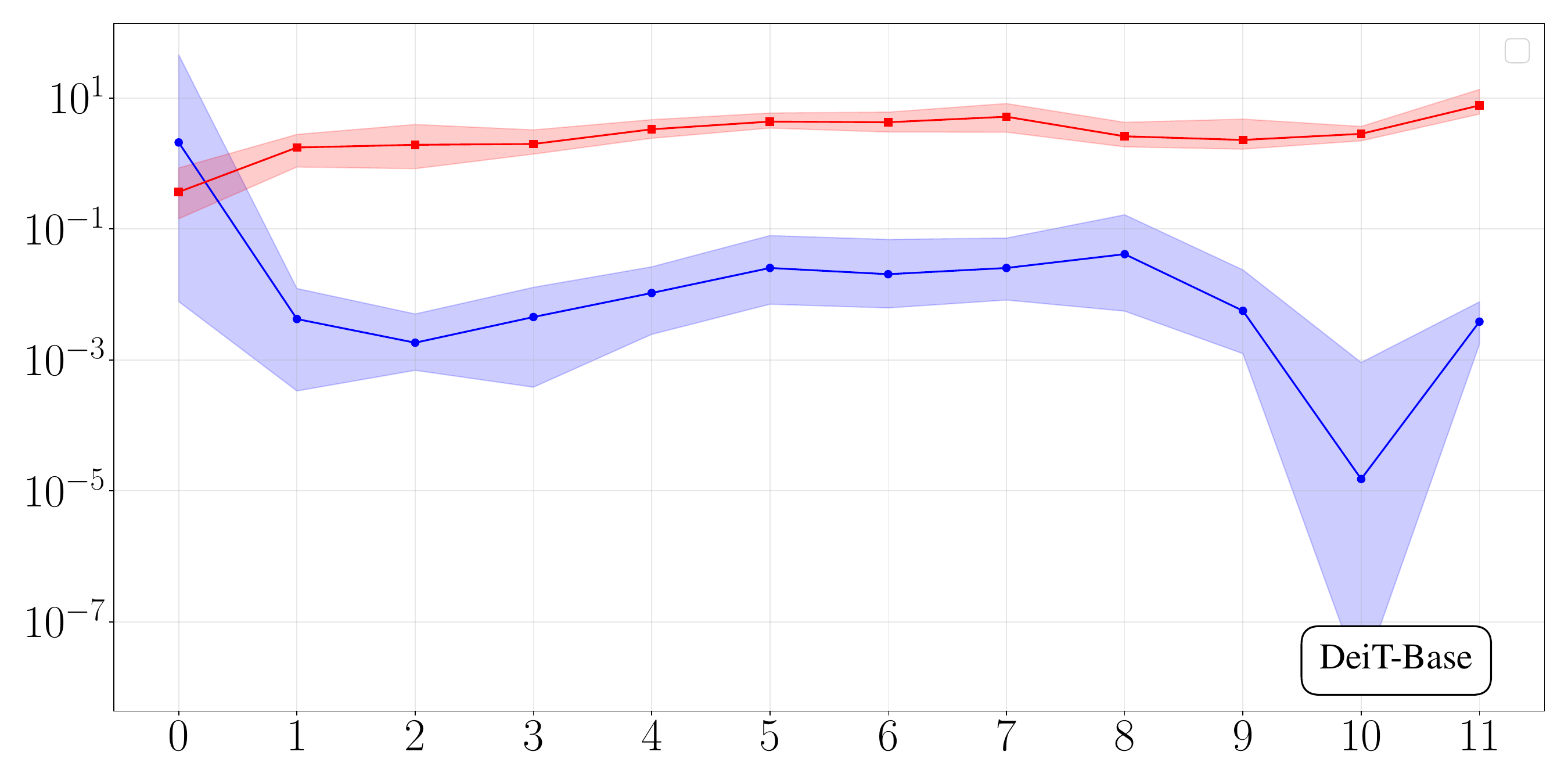} \\
        \includegraphics[width=0.33\textwidth]{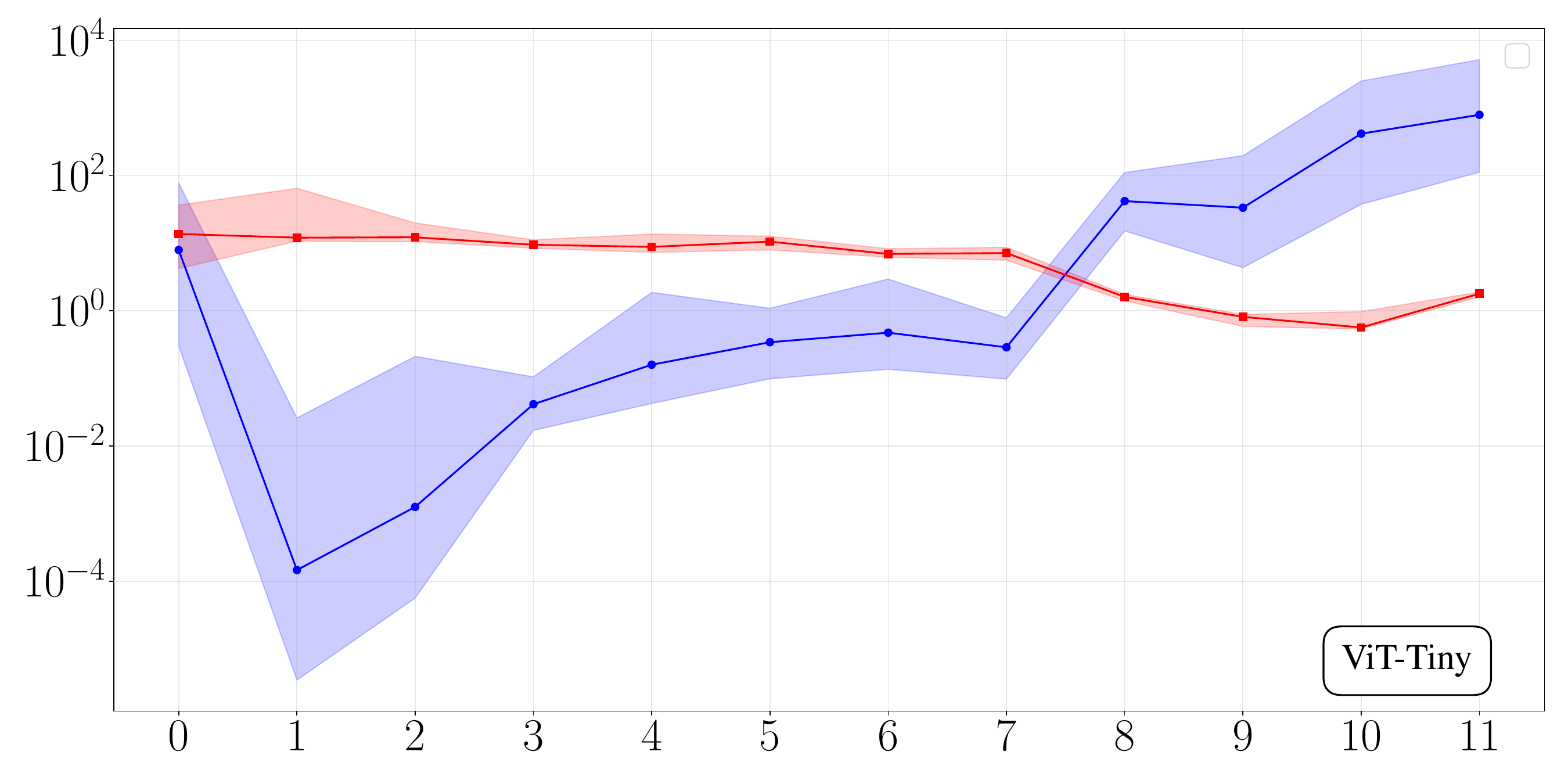} &
        \includegraphics[width=0.33\textwidth]{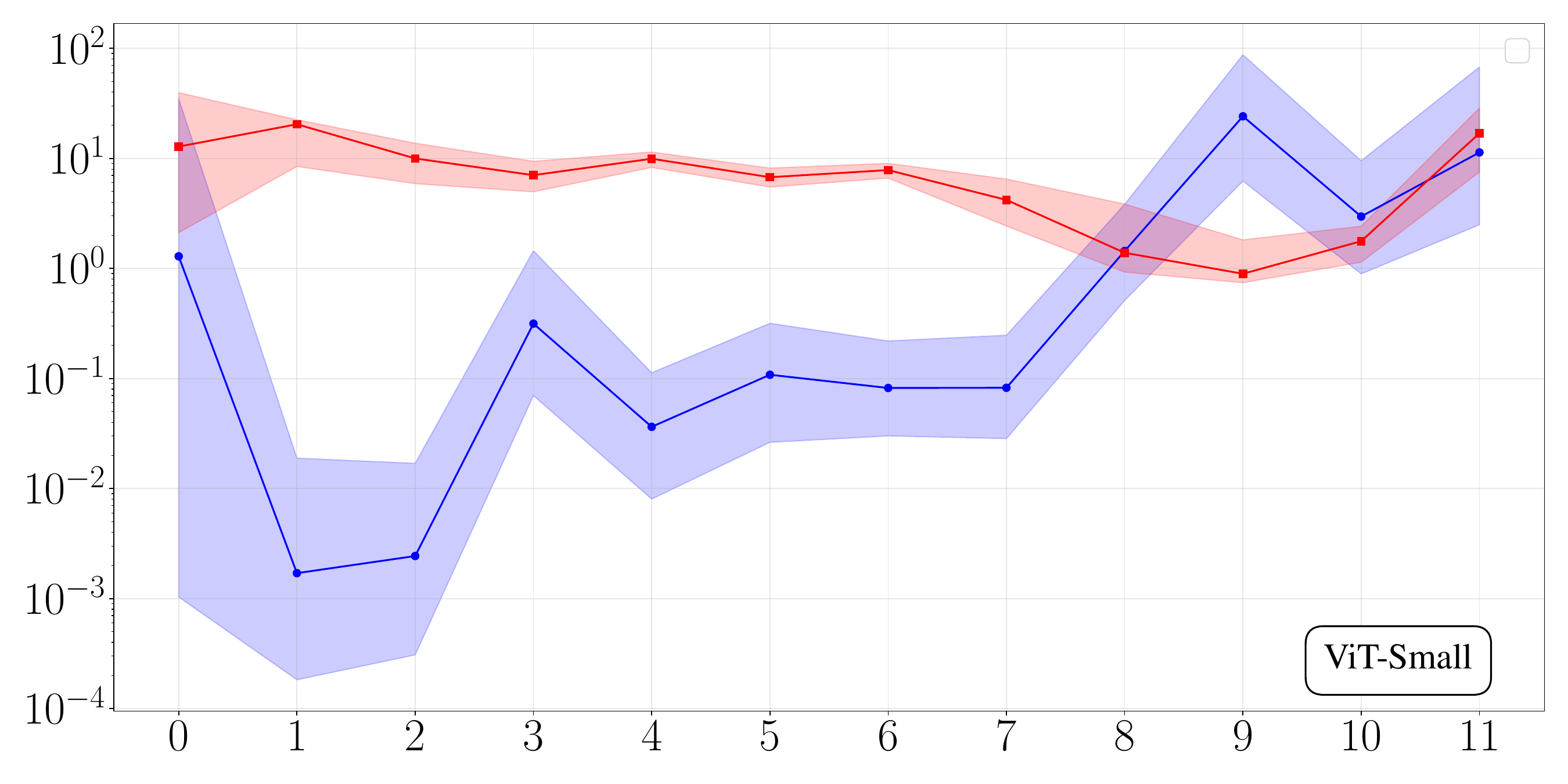} &
        \includegraphics[width=0.33\textwidth]{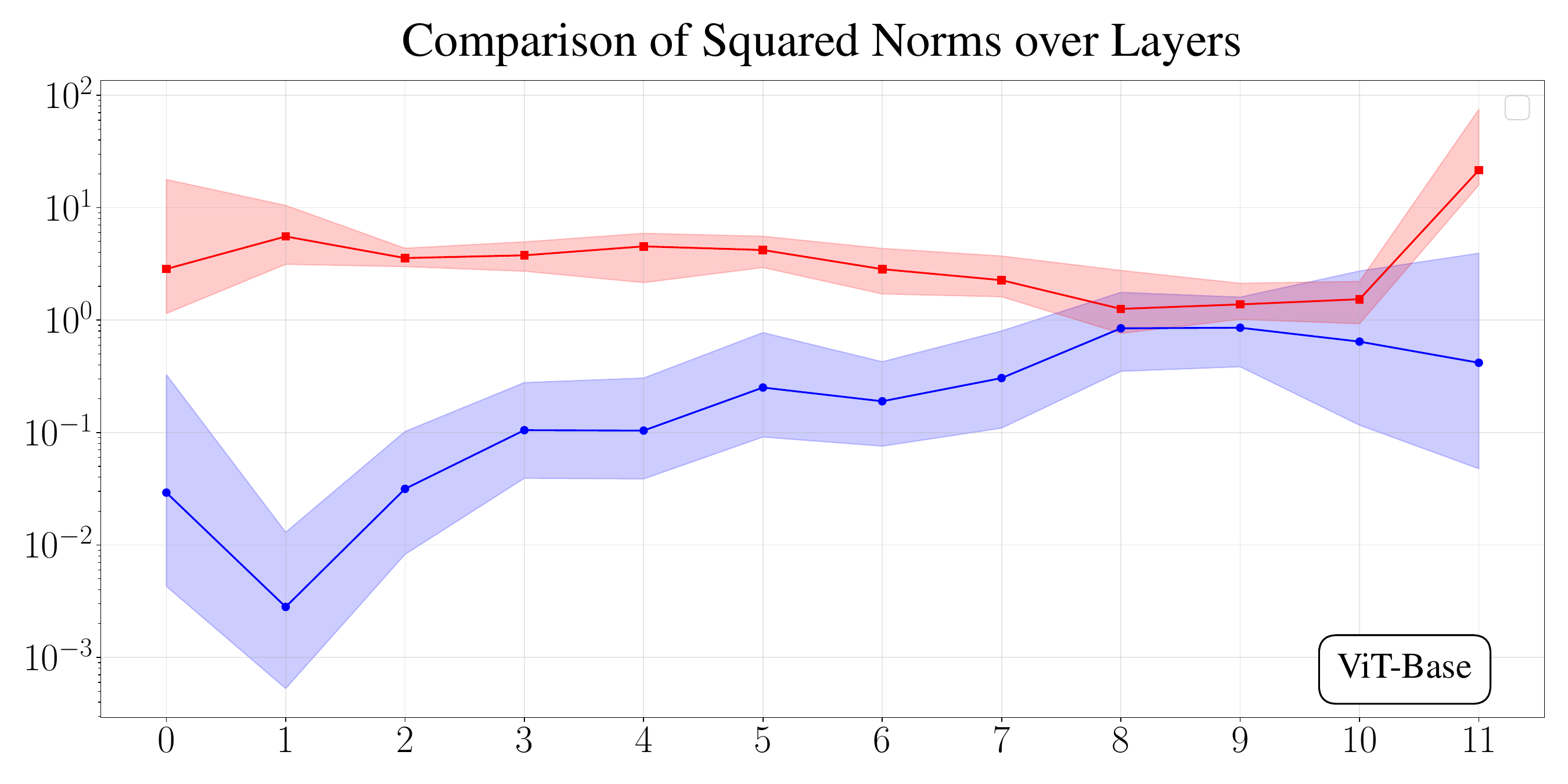} \\
    \end{tabular}
    \caption{Comparison of squared norms across transformer layers. The plots show medians (solid lines) and 95\% percentiles (shaded regions) of $\left\| \varphi(q_i) \right\|^2$ (blue) and $\left\| h_i \right\|^2$ (red) for 9 pre-trained transformer models for an input image. Values are displayed in log-scale due to the small magnitude of $\left\| \varphi(q_i) \right\|^2$. Log scaling highlights vanishing $\|\varphi(q_i)\|^2$ magnitudes. Notice the (1) high variance in $\varphi(q_i)$ projections vs. stable attention outputs, (2) no layer-wise convergence despite architectural scaling (DeiT/ViT, Tiny$\to$Base)}
\label{fig:norm_comparison}
\end{figure*}

The following plots in Figure~\ref{fig:norm_comparison} demonstrate that the theoretically calculated values of \( \| \varphi(q_i) \|^2 \), derived from a pretrained model, fail to align with the squared norms of the output vectors \( \| h_i \|^2 \) over different layers. While these measures occasionally converge to a similar scale in deeper layers, they remain distinct. 

\begin{figure}[H]
    \centering
    \includegraphics[width=1.1\linewidth]{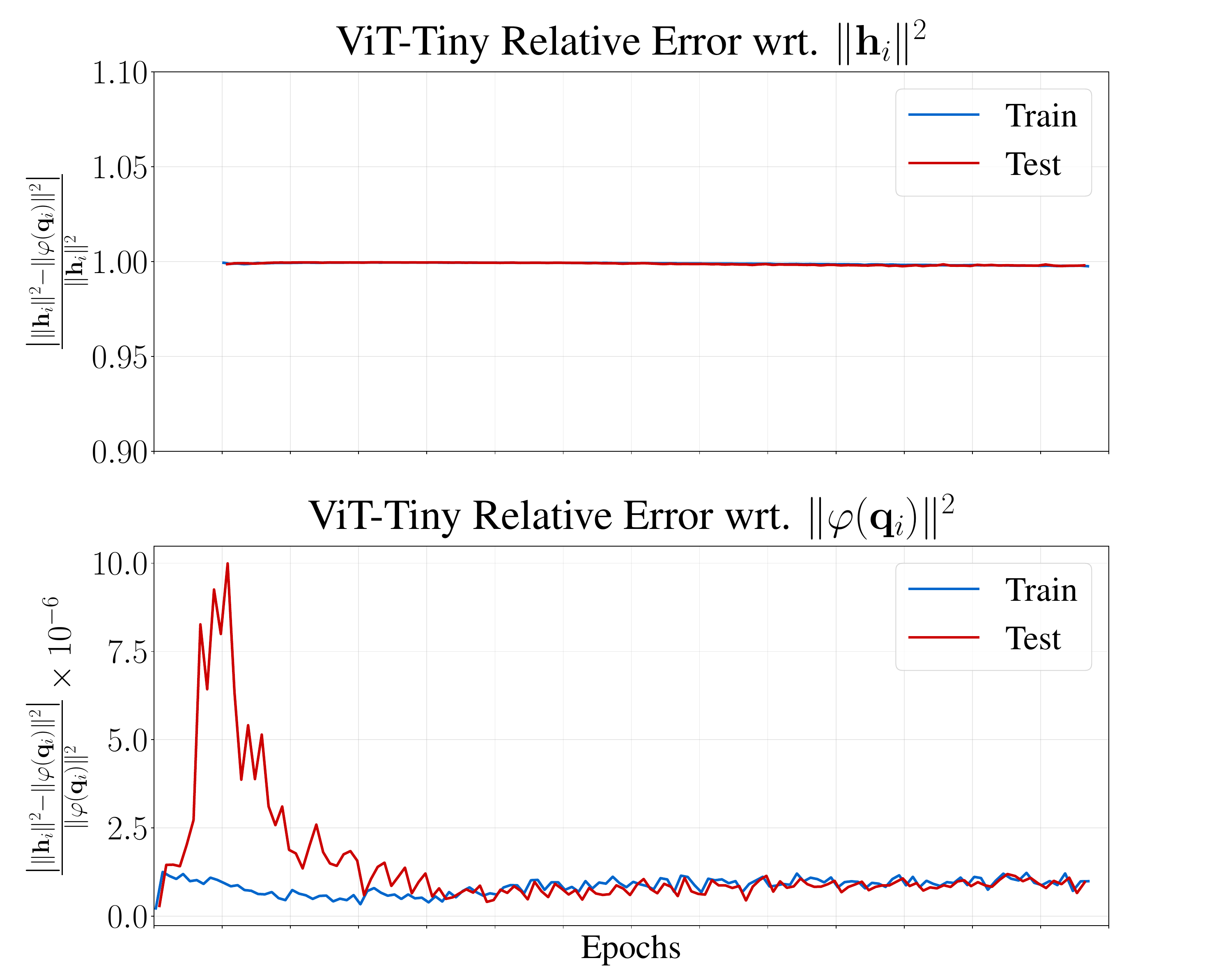}
    \caption{Relative absolute reconstruction train/test errors with respect to $\|\varphi(\mathbf{q}_i)\|^2$ and $\|h_i\|^2$ for ViT-Tiny. Errors with respect to $\|\varphi(\mathbf{q}_i)\|^2$ are in scale $10^{-6}$. For clarity, the lower panel excludes the first 10 epochs to mitigate outlier effects and enhance trend visibility.}
    \label{fig:min_error_plots}
\end{figure}

Plots in Figure~\ref{fig:min_error_plots} demonstrate that the relative error for $\|h_i\|^2$ remains near $1.0$ during training, while the error for $\|\varphi(q_i)\|^2$ spans $\sim\!10^6$—highlighting a stark magnitude disparity. This discrepancy underscores that the observed decrease in projection error $J_\text{proj}$ does not imply convergence as initially suggested.

\section{Eigenvalue Analysis}
\label{app:eigenvalue}
\begin{figure*}[t]
    \centering
    \begin{tabular}{@{}c@{\hspace{0.02\textwidth}}c@{}}
        \includegraphics[width=0.48\textwidth]{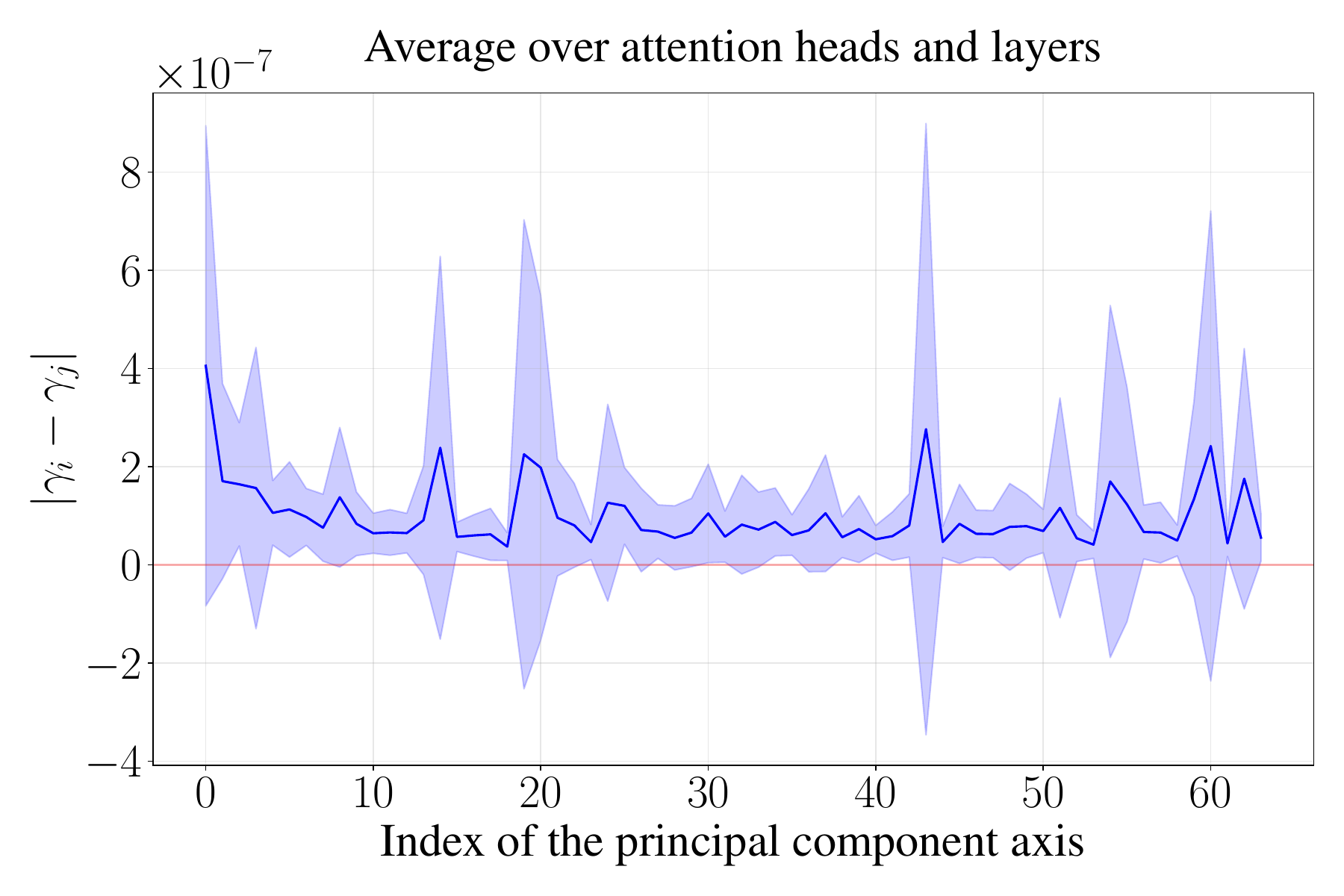} &
        \includegraphics[width=0.48\textwidth]{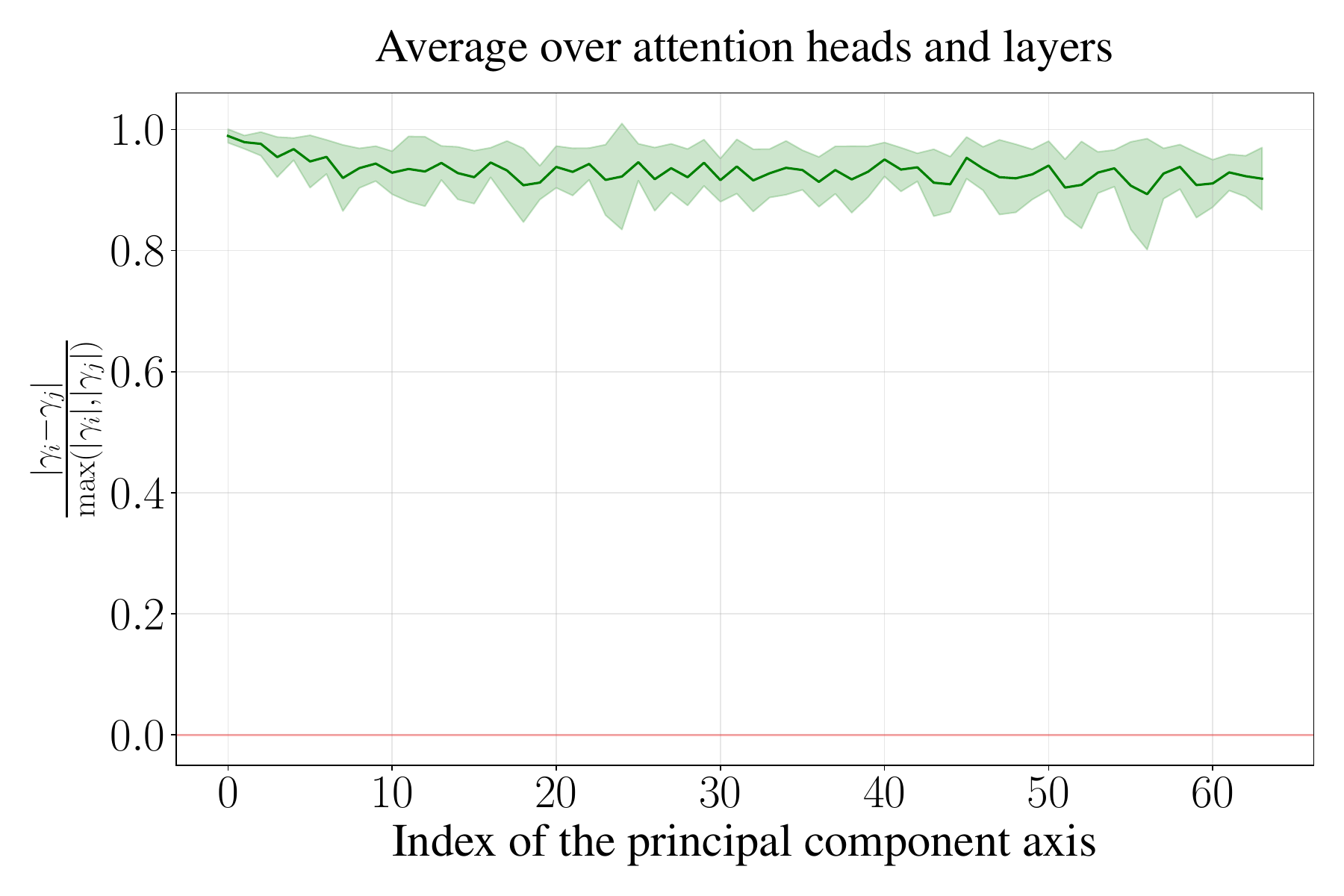} \\
        \includegraphics[width=0.48\textwidth]{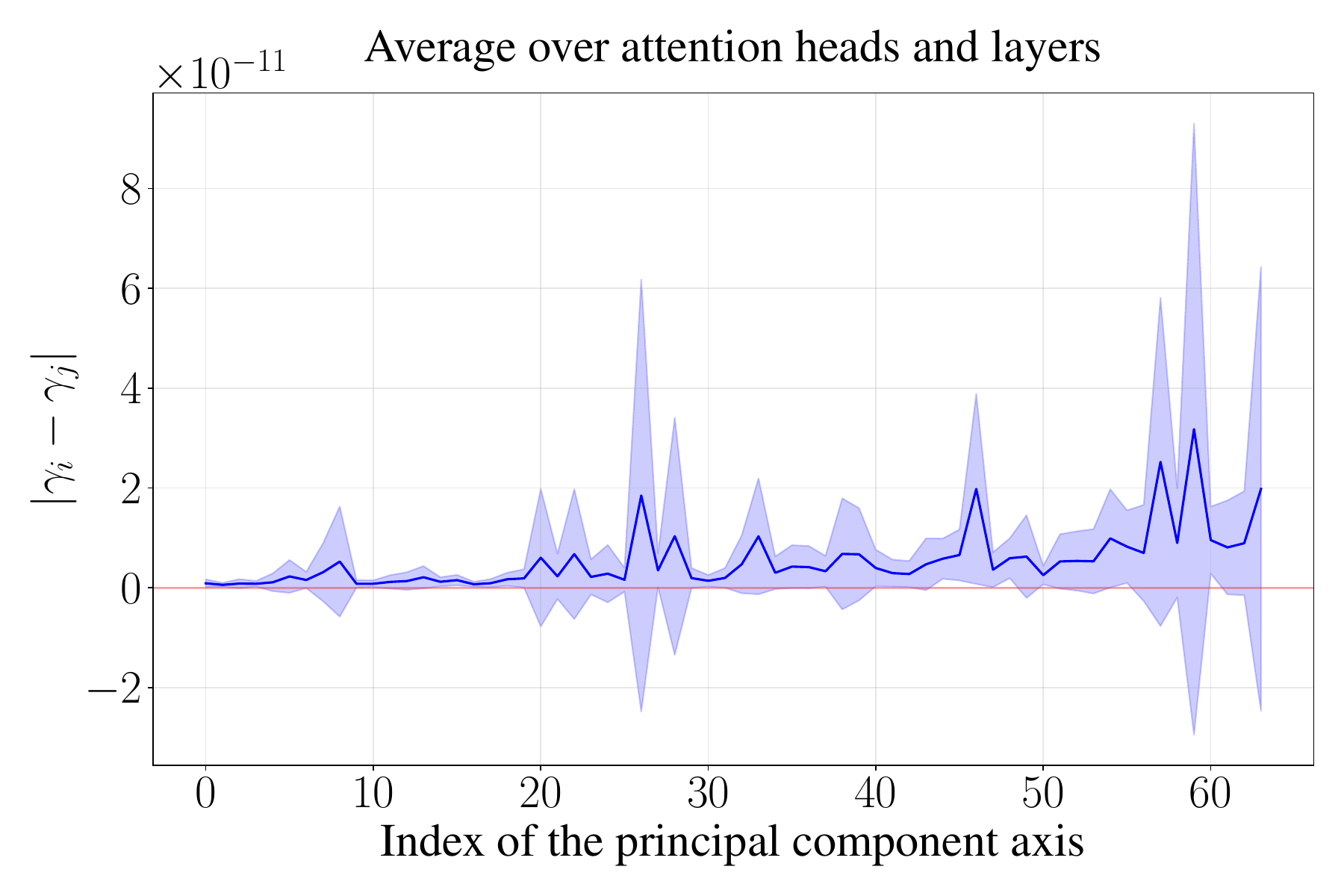} &
        \includegraphics[width=0.48\textwidth]{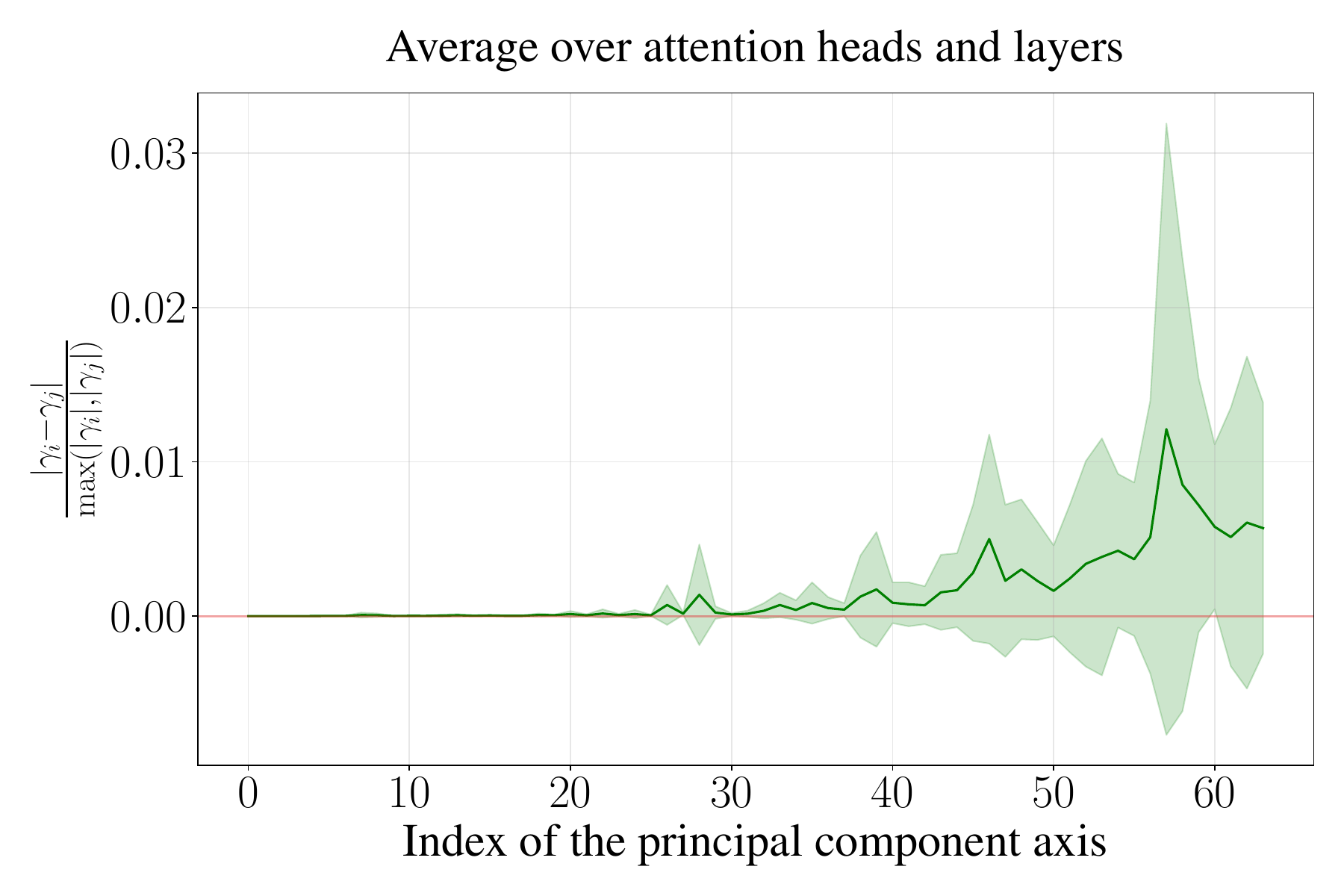} \\
    \end{tabular}
    \caption{ (ViT Tiny)
    Top row: Mean and standard deviation of the absolute differences of entries in the $\gamma$ vector from true eigenvectors of $\tilde{K}_{\varphi}$ (matrix $A$). 
    Bottom row: Corresponding results for random-direction eigenvectors ($A_{\text{random}}$) with matched row norms. 
    Left panels initially suggest both satisfy $\frac{\tilde{K}_{\varphi} \hat{a}_d}{N \hat{a}_d} = \gamma = [\gamma_1, \ldots, \gamma_N]$ with $\gamma_1 = \cdots = \gamma_N = \text{const.}$; 
    however, absolute differences reveal orders-of-magnitude deviation ($10^{-7}$ vs. $10^{-11}$). 
    Right panels (relative error to $\max(|\gamma_i|, |\gamma_j|)$) demonstrate the condition violation more explicitly through significantly higher relative errors for $A_{\text{random}}$, 
    showing small $\tilde{K}_{\varphi}$ eigenvalues permit visual resemblance despite failing the eigenvector criterion.
    }
\label{fig:gamma_comparison}
\end{figure*}

To empirically demonstrate that visualizations resembling the original authors' results can emerge even without strict adherence to the eigenvector condition $\tilde{K}_{\varphi} \hat{a}_d = \lambda \hat{a}_d$, we generate a perturbed matrix $A_\text{random}$ by adding standard Gaussian noise scaled by $0.1$ to each entry of $A$, followed by $QR$-decomposition to re-orthogonalize its columns. In Figure~\ref{fig:gamma_comparison}, we show two cases where the initial type of plots can be misleading, whereas the second plots reveal the difference between them.

\subsection{Eigenvalue Statistics Calculation}
\label{app:eig_stats}

For each image, we compute the absolute eigenvalues of the attention mechanism for every head and layer. These eigenvalues are grouped by their \textit{rank} (sorted position) across all heads and layers. We then compute the average eigenvalue value \textit{for each rank position} (e.g., the mean of all 1st-largest eigenvalues, the mean of all 2nd-largest eigenvalues, etc.). From these rank-wise averages, we calculate four statistics— max, min, mean, and median—across all rank positions. Finally, we report the mean and standard deviation of these statistics over 25 randomly sampled images from \imagenet. 

For certain transformer architectures, direct eigenvalue computation exhibited numerical instability due to floating-point precision limitations. We resolved this by standardizing key ($k$) vectors (i.e., subtracting means and dividing by standard deviations per dimension) prior to covariance matrix construction. While standardization during inference risks severely degrading model performance, Table~\ref{tab:app_eigenvalues_comparison} reveals that its impact on the eigenvalues of the centered Gram matrix $\tilde{K}_{\varphi}$ is negligible. This confirms that discrepancies with \cite{teo2024unveilinghiddenstructureselfattention} arise from undocumented methodological choices, not pre-processing steps.

\textbf{Relative projection error $J_\text{proj}$ plots reveal a fundamental flaw in the "reconstruction loss minimization" argument:} During the training, $\|\varphi(q_i)\|^2$ (bottom) remains negligible ($\sim 10^{-3}$) compared to $\|h_i\|^2$ (top). This disparity confirms that decreasing $J_\text{proj}$ arises not from alignment between $\varphi(q_i)$ and reconstructions, but from collapsing $\|h_i\|^2$ magnitudes. A similar inconsistency is observed for language models in Appendix~\ref{sec:language_models}.

\begin{table}[htbp]
\centering
\caption{Eigenvalue statistics with and without(*) query-key standardization ($\times 10^{-6}$)}
\label{tab:app_eigenvalues_comparison}
\scalebox{0.57}{
\begin{tabular}{@{}l|c|c|c|c@{}}
\hline
\hline
    \multirow{2}{*}{\textbf{Model}} & \multicolumn{4}{c}{\textbf{Eigenvalue Statistics}} \\ \cline{2-5} \\[-0.9em]
    & Max & Min & Mean & Median \\
    \midrule
    ViT-Tiny & $147 \pm 11$ & $17 \pm 5$ & $37 \pm 7$ & $30 \pm 7$ \\
    ViT-Tiny* & $178 \pm 19$ \qinc{21\%} & $4 \pm 2$ \qdec{76\%} & $16 \pm 3$ \qdec{57\%} & $9 \pm 3$ \qdec{70\%} \\
    \midrule
    ViT-Large & $177 \pm 22$ & $21 \pm 5$ & $42 \pm 6$ & $34 \pm 6$ \\
    ViT-Large* & $497 \pm 36$ \qinc{181\%} & $7 \pm 2$ \qdec{67\%} & $31 \pm 2$ \qdec{26\%} & $15 \pm 2$ \qdec{56\%} \\
    \midrule
    DeiT-Tiny & $325 \pm 5$ & $34 \pm 10$ & $65 \pm 10$ & $53 \pm 11$ \\
    DeiT-Tiny* & $1043 \pm 99$ \qinc{221\%} & $34 \pm 11$ \qinc{0\%} & $96 \pm 15$ \qinc{48\%} & $60 \pm 14$ \qinc{13\%} \\
    \midrule
    DeiT-Small & $306 \pm 4$ & $34 \pm 9$ & $66 \pm 11$ & $54 \pm 11$ \\
    DeiT-Small* & $1343 \pm 175$ \qinc{339\%} & $25 \pm 8$ \qdec{26\%} & $87 \pm 11$ \qinc{32\%} & $46 \pm 11$ \qdec{15\%} \\
    \midrule
    DeiT-Tiny-D & $205 \pm 7$ & $32 \pm 9$ & $61 \pm 10$ & $51 \pm 10$ \\
    DeiT-Tiny-D* & $796 \pm 95$ \qinc{288\%} & $27 \pm 8$ \qdec{16\%} & $78 \pm 12$ \qinc{28\%} & $49 \pm 11$ \qdec{4\%} \\
    \midrule
    DeiT-Small-D & $224 \pm 7$ & $33 \pm 9$ & $63 \pm 10$ & $53 \pm 10$ \\
    DeiT-Small-D* & $1153 \pm 151$ \qinc{415\%} & $23 \pm 7$ \qdec{30\%} & $78 \pm 10$ \qinc{24\%} & $43 \pm 10$ \qdec{19\%} \\
\hline
\hline
\end{tabular}}
\end{table}

\newpage
\subsection{Gram Matrix Eigenvalue Equation}
\label{app:derive}

In this subsection, we will derive the gram matrix eigenvalue equation explicitly. We begin with the following expressions:
\begin{align*}
u_d &= \sum_{j=1}^N a_{dj} \tilde{\varphi}(k_j) \\
\frac{1}{N}\sum_{j=1}^N& \tilde{\varphi}(k_j) \{ \tilde{\varphi}(k_j)^\top u_d \} = \lambda_d u_d
\end{align*}

\begin{figure*}
\text{We will be using the following matrix multiplication:}
\begin{align*}
    \tilde{K}_\varphi a_d = \begin{bmatrix} 
\tilde{\varphi}(k_1)^\top \tilde{\varphi}(k_1) & \tilde{\varphi}(k_1)^\top \tilde{\varphi}(k_2) & \cdots & \tilde{\varphi}(k_1)^\top \tilde{\varphi}(k_N) \\
\vdots & \vdots & \ddots & \vdots \\
\tilde{\varphi}(k_N)^\top \tilde{\varphi}(k_1) & \tilde{\varphi}(k_N)^\top \tilde{\varphi}(k_2) & \cdots & \tilde{\varphi}(k_N)^\top \tilde{\varphi}(k_N)
\end{bmatrix} &
\begin{bmatrix} 
a_{d1} \\ a_{d2} \\ \vdots \\ a_{dN} 
\end{bmatrix} 
\end{align*}
\text{whose entries can be expressed as:} $$
(\tilde{K}_\varphi a_d)_i = \sum_{j=1}^N \tilde{\varphi}(k_i)^\top \tilde{\varphi}(k_j) a_{dj} $$

Substituting $u_d$ as the weighted combination of $\tilde{\varphi}(k_j)$ yields: 
\begin{align*}
\frac{1}{N} \sum_{j=1}^{N} \tilde{\varphi}(k_j)\tilde{\varphi}(k_j)^{\top} \sum_{j'=1}^{N} a_{dj'}\tilde{\varphi}(k_{j'}) &= \lambda_d \sum_{j=1}^{N} a_{dj}\tilde{\varphi}(k_j) \\
\frac{1}{N} \sum_{j=1}^{N} \tilde{\varphi}(k_j) \underbrace{\sum_{j'=1}^{N} \tilde{\varphi}(k_j)^{\top} a_{dj'}\tilde{\varphi}(k_{j'})}_{(\tilde{K}_\varphi a_d)_j} &= \lambda_d \sum_{j=1}^{N} a_{dj}\tilde{\varphi}(k_j) \\
\frac{1}{N} \sum_{j=1}^{N} \tilde{\varphi}(k_i)^\top \tilde{\varphi}(k_j) \underbrace{\sum_{j'=1}^{N} \tilde{\varphi}(k_j)^{\top} a_{dj'}\tilde{\varphi}(k_{j'})}_{(\tilde{K}_\varphi a_d)_j} &= \lambda_d \underbrace{\sum_{j=1}^{N} \tilde{\varphi}(k_i)^\top a_{dj}\tilde{\varphi}(k_j)}_{(\tilde{K}_\varphi a_d)_i} \\
\frac{1}{N} \underbrace{\sum_{j=1}^{N} \tilde{\varphi}(k_i)^\top \tilde{\varphi}(k_j) (\tilde{K}_\varphi a_d)_j}_{(\tilde{K}_\varphi (\tilde{K}_\varphi a_d))_i} &= \lambda_d (\tilde{K}_\varphi a_d)_i \\
(\tilde{K}_\varphi (\tilde{K}_\varphi a_d))_i &= N\lambda_d (\tilde{K}_\varphi a_d)_i \\
\tilde{K}_\varphi \tilde{K}_\varphi a_d &= N\lambda_d \tilde{K}_\varphi a_d \\
\tilde{K}_\varphi (\tilde{K}_\varphi a_d - N\lambda_d a_d) &= 0
\end{align*}
\label{fig:eigenvector-derivation}
\end{figure*}

When $\tilde{K}_\varphi$ is invertible, the only solution is:
\begin{align*}
\tilde{K}_\varphi a_d = N\lambda_d a_d    
\end{align*}
which corresponds to the eigenvalue solution, where $a_d$ are eigenvectors of $\tilde{K}_\varphi$ with corresponding eigenvalues $N\lambda_d$.

If $\tilde{K}_\varphi$ is singular, additional solutions exist in the form $\{a_d | \tilde{K}_\varphi a_d - N\lambda_d a_d \in \text{Null}(\tilde{K}_\varphi)\}$. However, since the Gram matrix is symmetric and positive semi-definite, it can only be singular if it has a zero eigenvalue. In practice, using 10 different transformer models in our experiments shows that $\tilde{K}_\varphi$ is typically invertible, allowing us to assume that the solutions $a_d$ are eigenvectors.

\newpage
\section{Language Models}
\label{sec:language_models}
\begin{figure*}[t]
    \centering Distribution of
    $\left\| \varphi(q_i) \right\|^2$ (blue) and $\left\| h_i \right\|^2$ (red) across transformer layers (log-scale)
    \begin{tabular}{@{}c@{\hspace{0.02\textwidth}}c@{\hspace{0.02\textwidth}}c@{}}
        \includegraphics[width=0.33\textwidth]{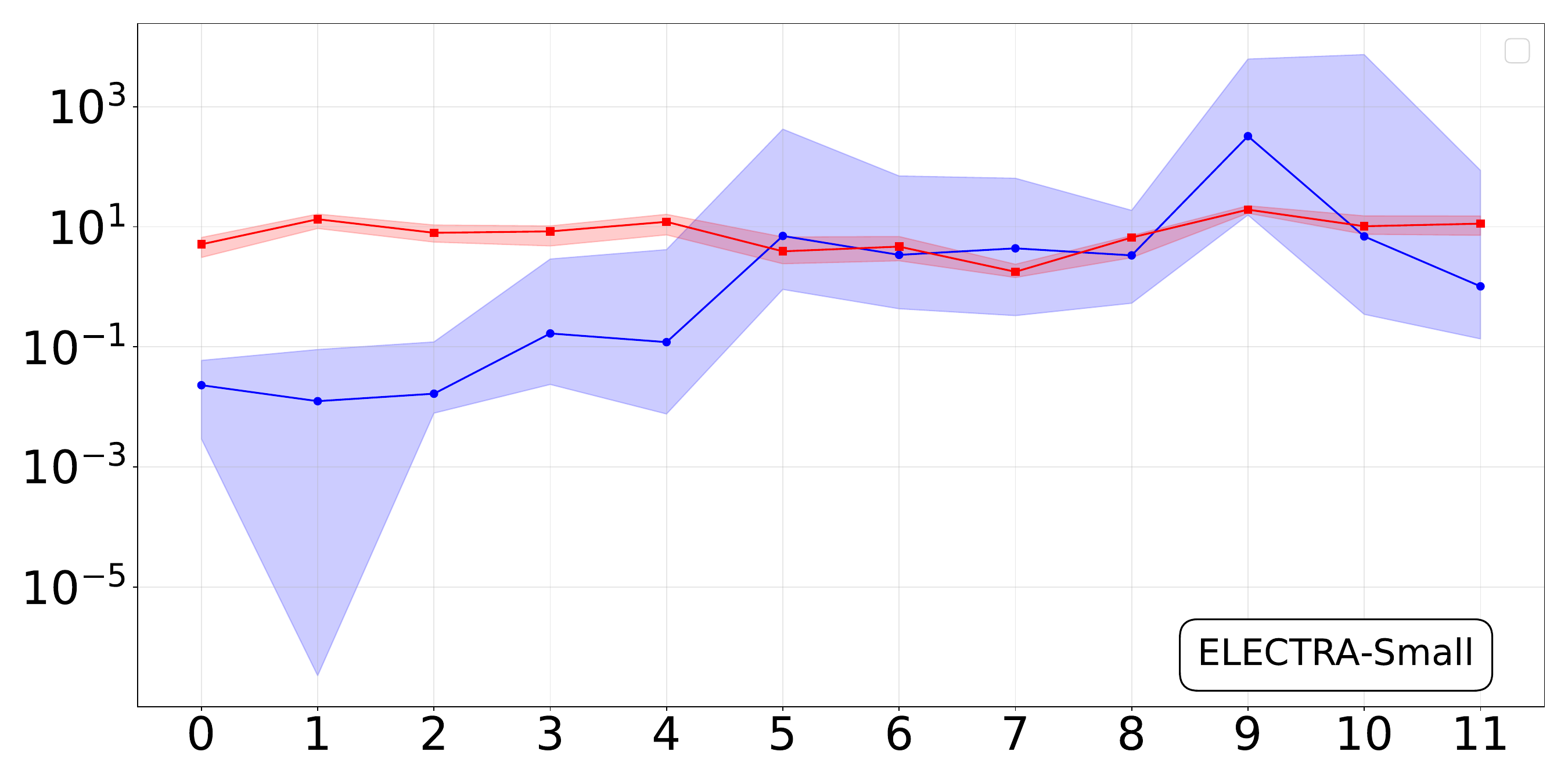} &
        \includegraphics[width=0.33\textwidth]{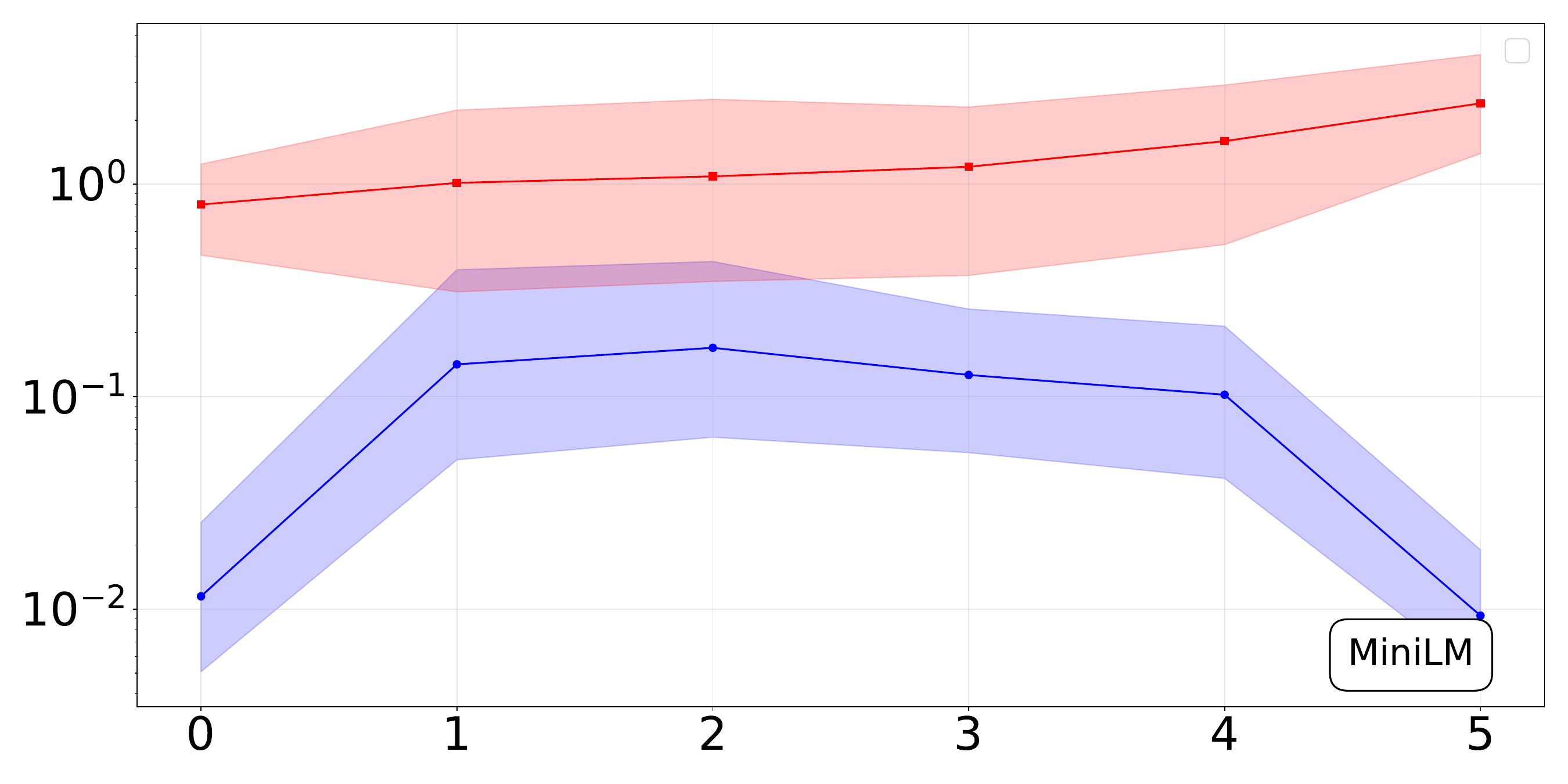} &
        \includegraphics[width=0.33\textwidth]{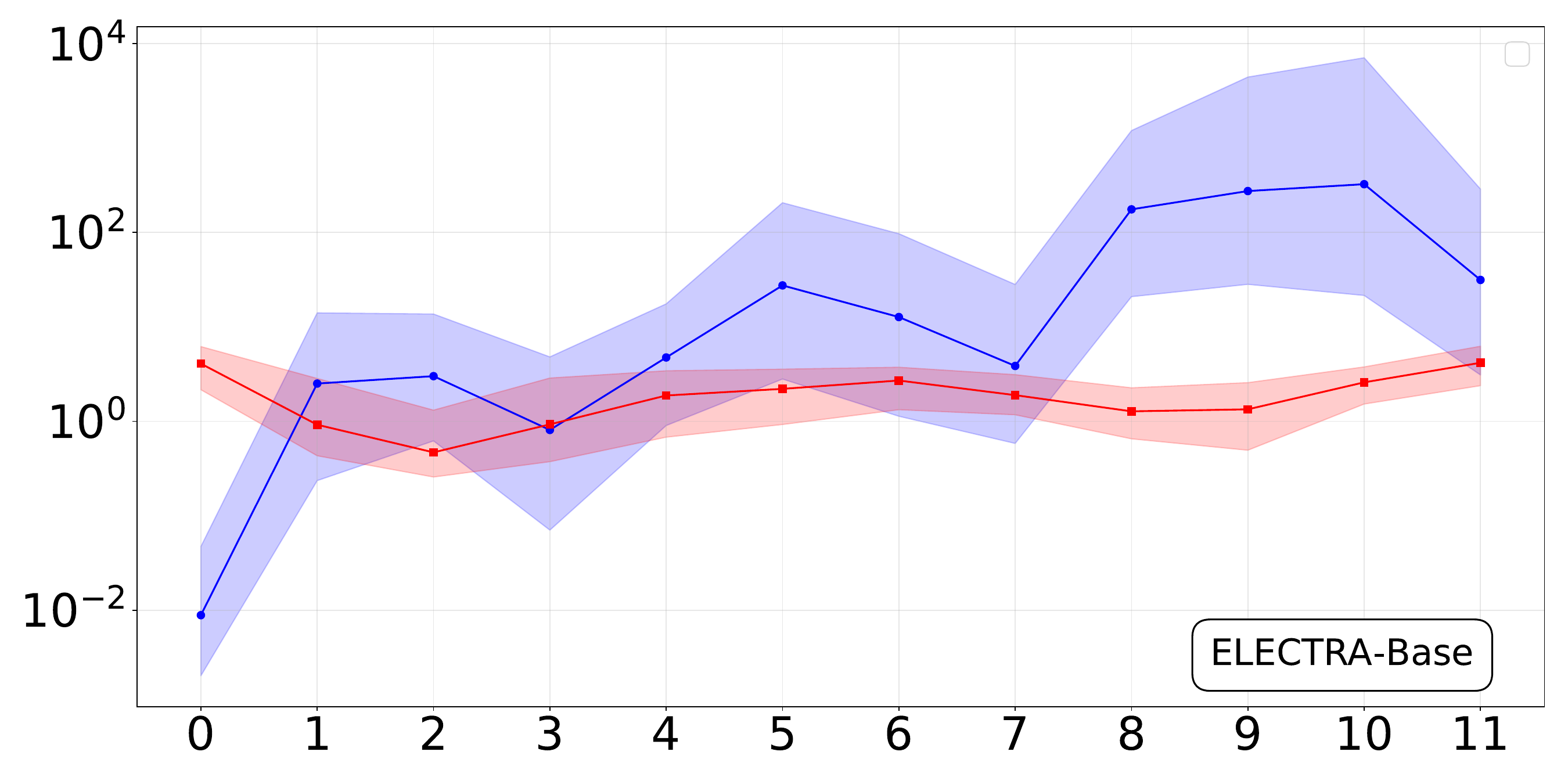} \\
        \includegraphics[width=0.33\textwidth]{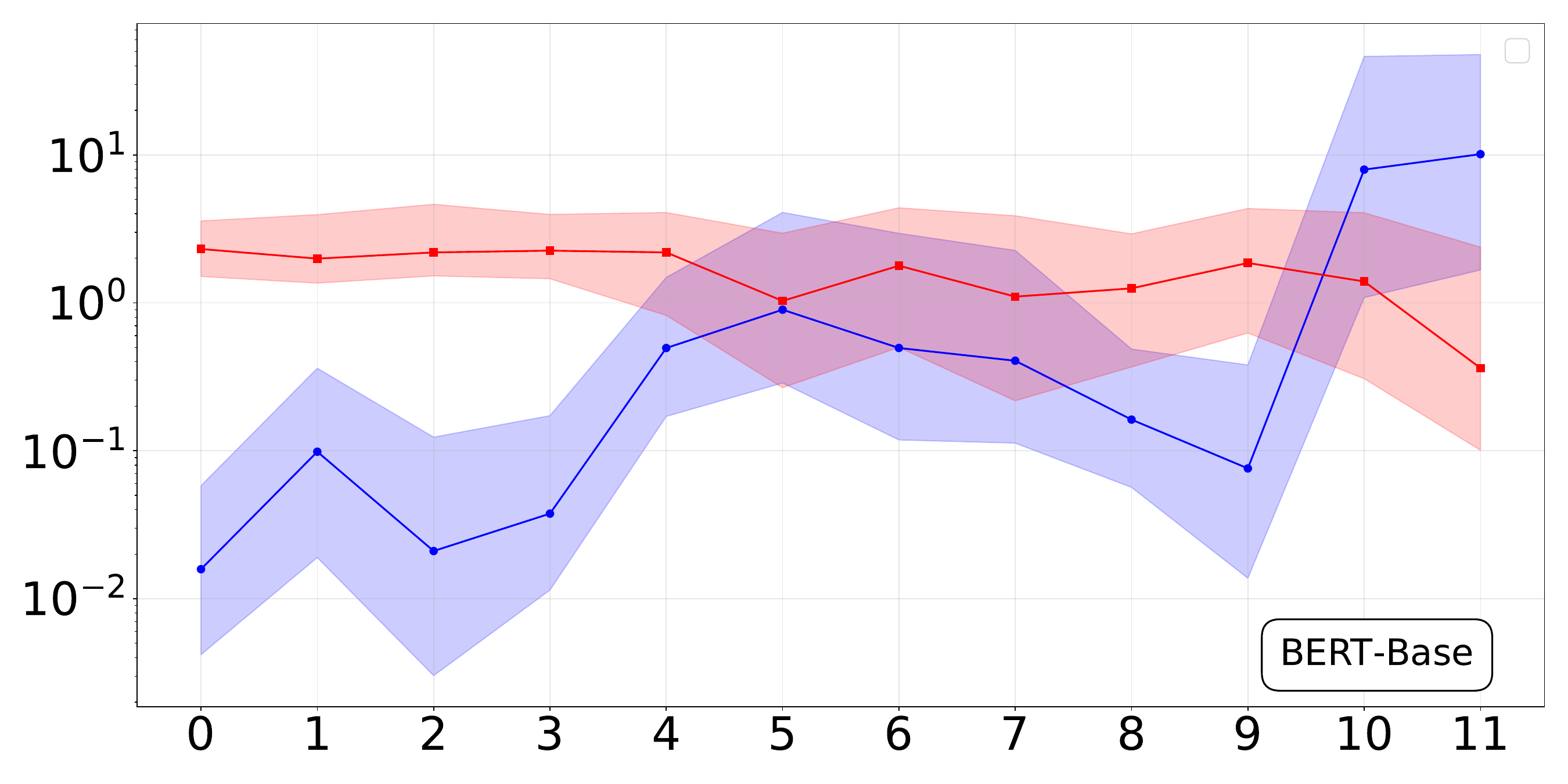} &
        \includegraphics[width=0.33\textwidth]{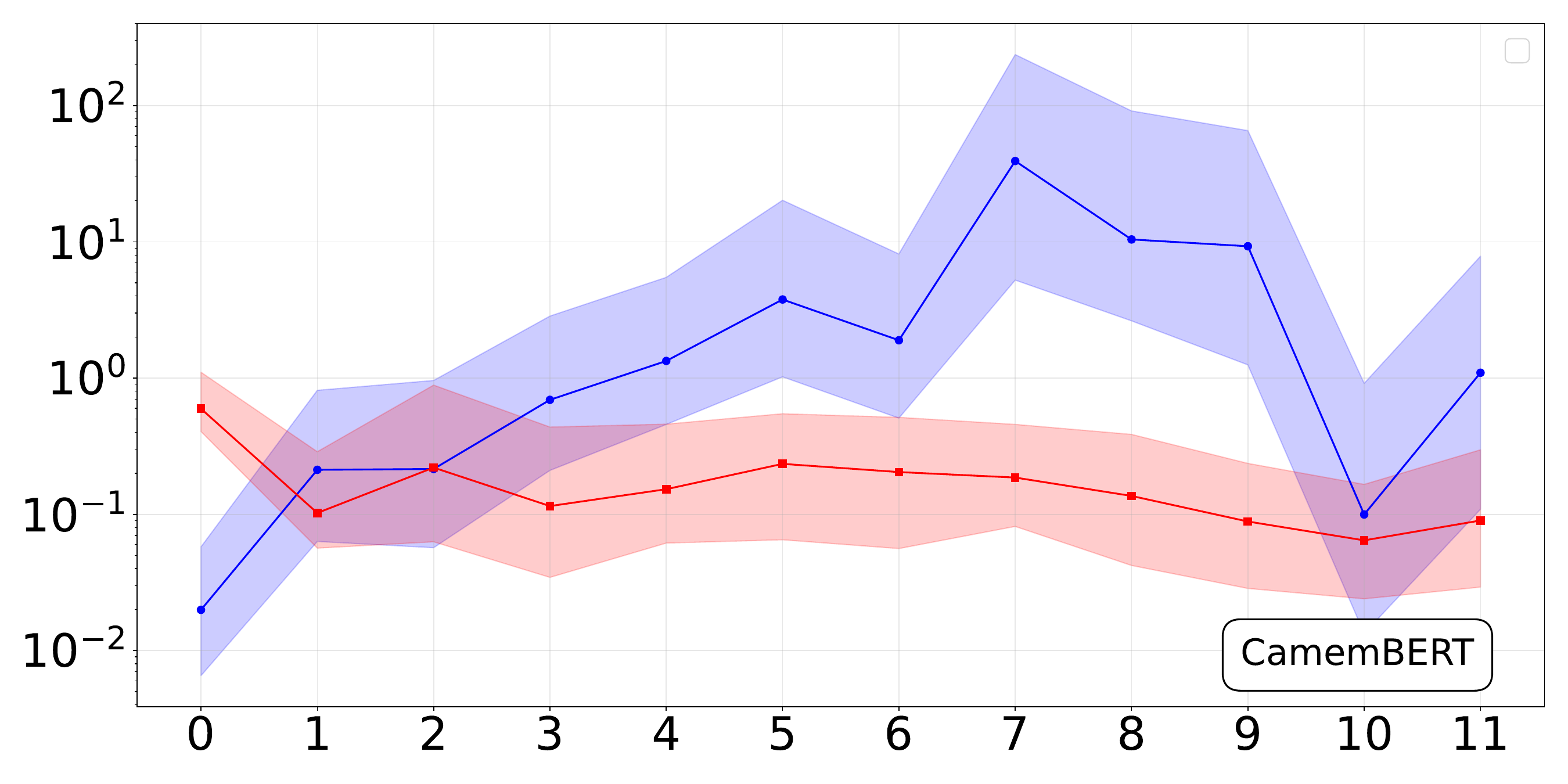} &
        \includegraphics[width=0.33\textwidth]{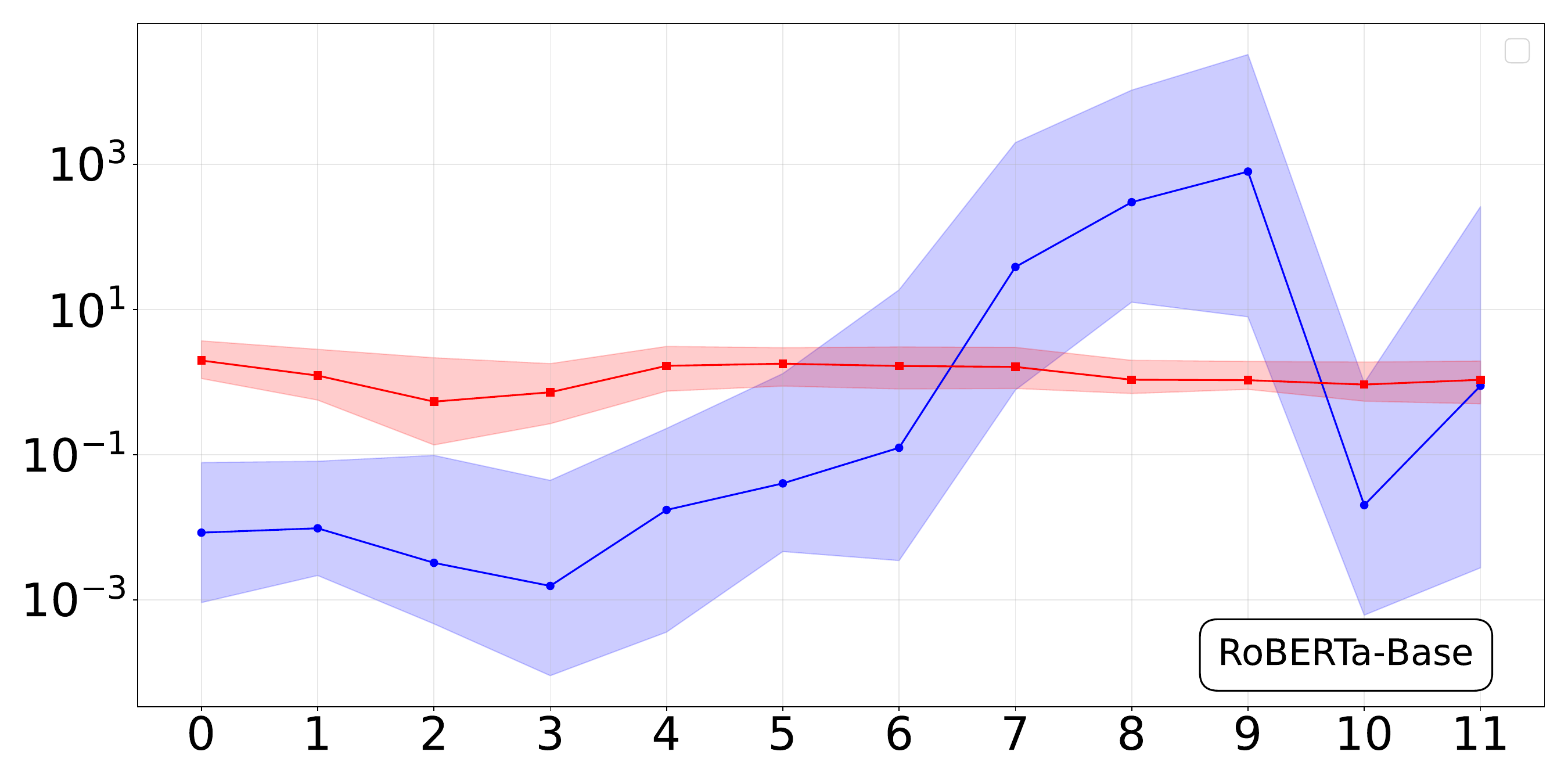} \\
        \includegraphics[width=0.33\textwidth]{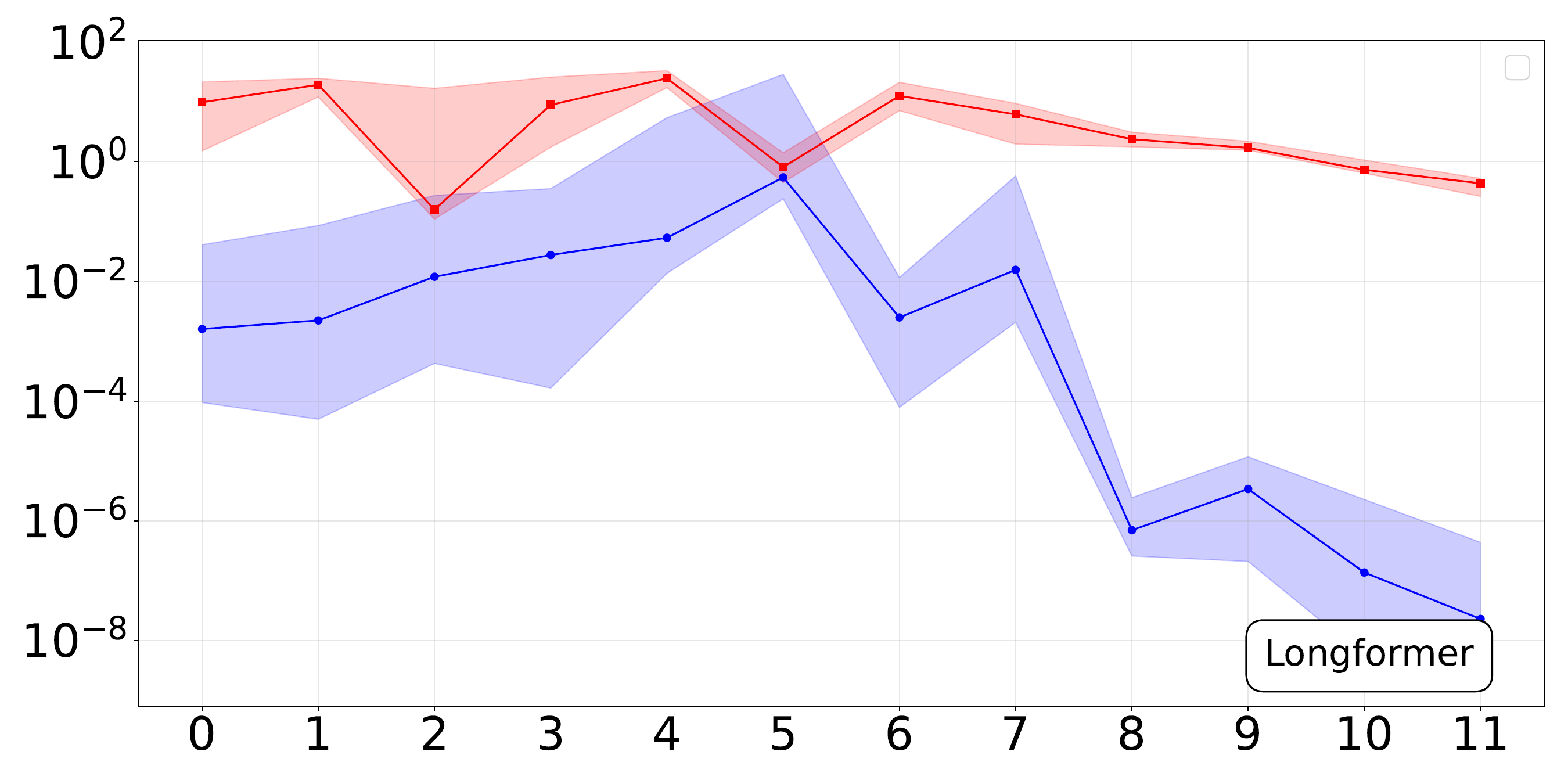} &

        \includegraphics[width=0.33\textwidth]{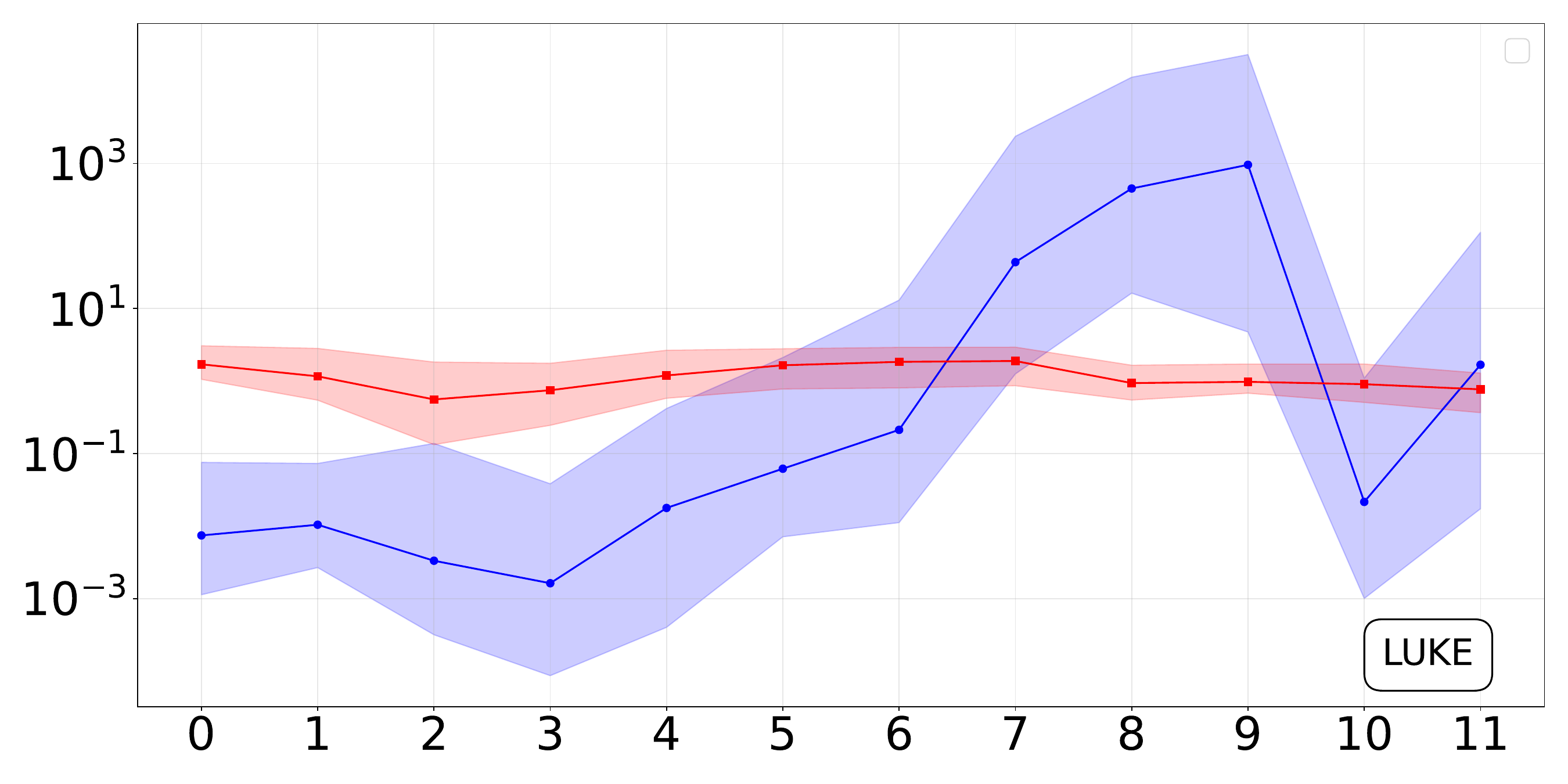} &
        \includegraphics[width=0.33\textwidth]{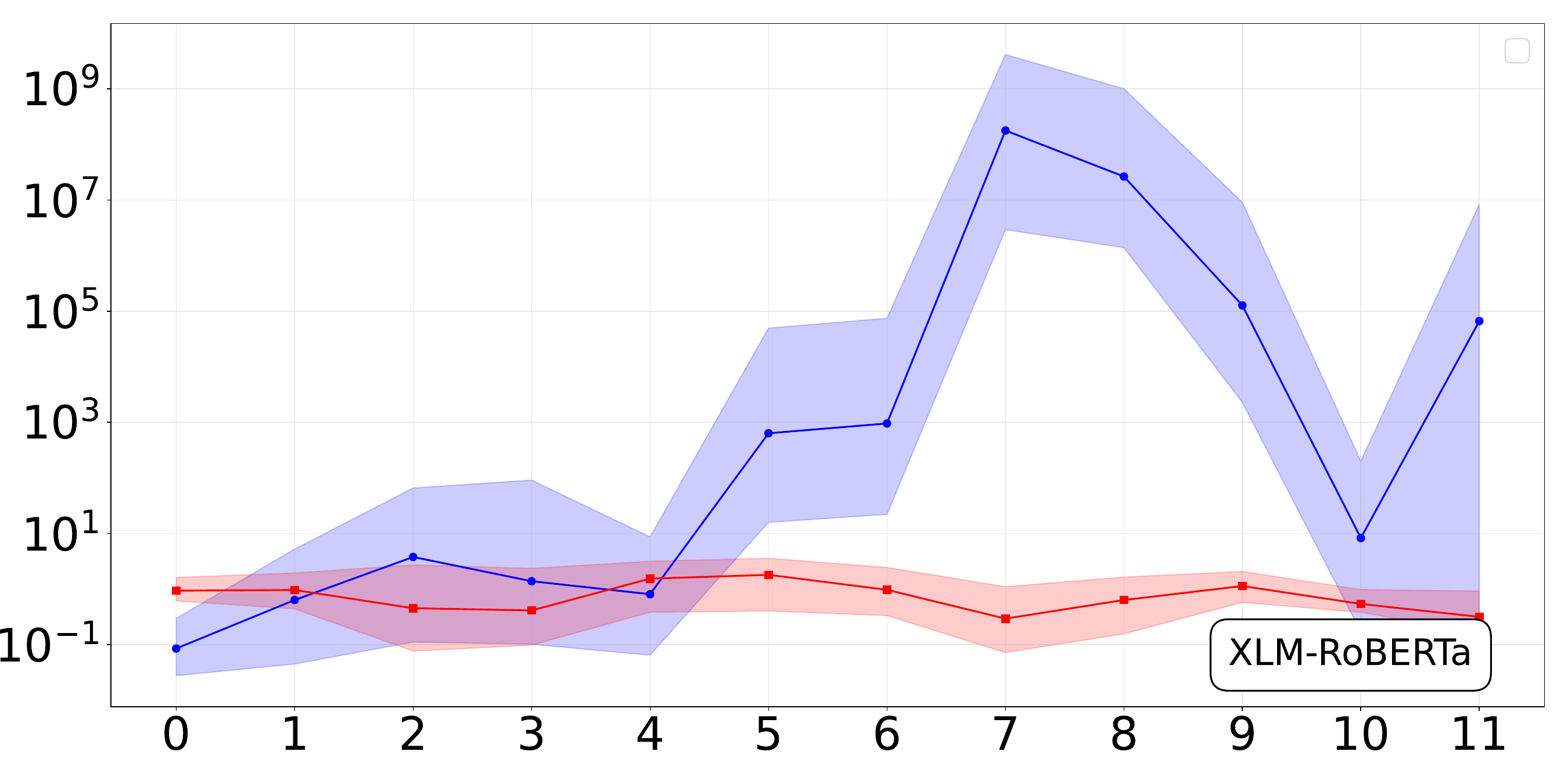} \\
    \end{tabular}
    \caption{ Distribution of $\left| \varphi(q_i) \right|^2$ (blue) and $\left| h_i \right|^2$ (red) across layers of nine pre-trained encoder-only language models (log scale) (ordered by the parameter count).
Each plot shows the median (solid line) and 95th percentile (shaded region) of the squared norm values across tokens. Despite differences in architecture and scale, all models exhibit a similar pattern: large variability in $\left| \varphi(q_i) \right|^2$ compared to the more stable $\left| h_i \right|^2$, and no consistent convergence behavior across layers. This mirrors observations made in vision transformers. }
\label{fig:nlp_norm_comparison}
\end{figure*}

Same experiments on encoder-only language models in Figure~\ref{fig:nlp_norm_comparison} reveals a similar pattern. As our models, we utilized \texttt{bert-base-uncased} \cite{DBLP:journals/corr/abs-1810-04805}, \texttt{roberta-base} \cite{DBLP:journals/corr/abs-1907-11692}, \texttt{electra-small-discriminator} , \texttt{electra-base-discriminator} \cite{clark2020electrapretrainingtextencoders}, \texttt{xlm-roberta-base} \cite{DBLP:journals/corr/abs-1911-02116}, \texttt{longformer-base-4096} \cite{Beltagy2020Longformer}, \texttt{all-MiniLM-L6-v2} \cite{reimers-2019-sentence-bert}, \texttt{camembert-base} \cite{martin2020camembert}, \texttt{luke-base} \cite{yamada2020luke}.

Table~\ref{tab:nlp_similarity_measures} reveals that, across all examined language‑model encoders, the similarity between the attention‑learned value matrix $V$ and its KPCA‑based approximation $\dot{V}_\text{KPCA}$ remains disappointingly low, indicating that the proposed reconstruction is no more effective for NLP models than for their vision counterparts. We used 100 randomly sampled images from WikiText-103 dataset \cite{merity2016pointer}.

\begin{table}[ht]
    \centering
    \caption{Similarity results between attention‑learned value matrix $V$ and proposed $\dot{V}_\text{KPCA}$ on a range of NLP encoder models.  
    MDC: Max Direct Cosine Similarity; MOC: Max Optimal Cosine Similarity (Jonker–Volgenant matching); LCKA: Linear CKA; KCKA: Kernel CKA.  
    Models are listed from the smallest to the largest (approximate) parameter count.}
    \label{tab:nlp_similarity_measures}
    \scalebox{0.85}{\begin{tabular}{l|c|c|c|c}
\hline
\hline
    \multirow{2}{*}{\;\;\;\;\;\textbf{Model}} & \multicolumn{4}{c}{\textbf{Similarity Measures}} \\ \cline{2-5} \\[-0.9em]
    & MDC & MOC & LCKA & KCKA \\
    \midrule
    ELECTRA‑Small & 0.22 & 0.40 & 0.07 & 0.28 \\
    MiniLM & 0.40 & 0.57 & 0.13 & 0.38 \\
    BERT‑Base & 0.30 & 0.45 & 0.07 & 0.29 \\
    CamemBERT & 0.30 & 0.46 & 0.09 & 0.30 \\
    ELECTRA‑Base & 0.27 & 0.46 & 0.05 & 0.29 \\
    RoBERTa‑Base & 0.15 & 0.30 & 0.05 & 0.35 \\
    Longformer & 0.18 & 0.21 & 0.03 & 0.45 \\
    LUKE & 0.14 & 0.30 & 0.05 & 0.34 \\
    XLM‑RoBERTa & 0.20 & 0.38 & 0.05 & 0.29 \\
\hline
\hline
    \end{tabular}}
\end{table}

\section{Final Comments}
\label{sec:final_comments}

\noindent
\paragraph{KSVD \emph{v.} KPCA perspectives}  
While both KPCA and Kernel SVD (KSVD) interpret self‑attention through kernel methods, they differ fundamentally in what they \emph{guarantee}.  
The KPCA view of \citet{teo2024unveilinghiddenstructureselfattention} claims that the canonical mechanism \emph{by itself} drives the value matrix \(V\) towards the eigenvectors of the centred Gram matrix of the keys - which \textit{fails} under our empirical scrutiny.  
In contrast, the KSVD formulation of \citet{chen2023primalattentionselfattentionasymmetrickernel} finds a \textit{resemblance} between vanilla self‑attention output with the \emph{dual} representation of an asymmetric‐kernel SVD and makes no claim of spontaneous convergence.

\medskip
\noindent
\paragraph{The additional KSVD regulariser}  
To realize the KSVD in practice, \citet{chen2023primalattentionselfattentionasymmetrickernel} augment the task loss with a variance‑maximisation term
\begin{align}
  \label{eq:ksvd_obj}
  \min_{\Theta}\;
     \mathcal{L}_{\text{task}}(\Theta)\;+\;
     \eta \sum_{l=1}^{L}\!\!J_l
\end{align}
where \(J_l\) is the KSVD loss of the \(l\)-th Primal‑Attention layer, averaged over heads, and \(\eta>0\) is a hyper‑parameter.%
\footnote{See Eqs.\,(6–7) in \citet{chen2023primalattentionselfattentionasymmetrickernel} for the exact form of \(J_l\).}
Solving \eqref{eq:ksvd_obj} forces the dual variables
\(\{h_{rj}\}_{j=1}^{N}\) in
\(
e(x_i)=\sum_{j=1}^{N} h_{rj}\, \kappa(x_i,x_j)
\)
to become orthonormal right singular vectors of the asymmetric kernel matrix
\(K_{ij}=\kappa(x_i,x_j)\) \citep{SUYKENS2016600}.  

\medskip
\noindent
\textbf{Implication for canonical self‑attention}  
Without the regulariser (\(\eta=0\)) canonical self‑attention provides at most an \emph{interpretive lens}: the value vectors can be \emph{identified algebraically} with some set of dual coefficients, but they are \emph{not} guaranteed to align with the principal right singular directions of \(K\).  
Such alignment – and the attendant orthogonality/variance properties – emerges solely after optimising the joint objective~\eqref{eq:ksvd_obj}.  Hence, unlike the strong convergence asserted under the KPCA view, the KSVD lens remains descriptive unless that additional constraint is enforced during training.

\end{document}